\documentclass{article}
\usepackage[%
journal=MSL,    
lang=british,   
]{ems-journal}

\usepackage{mathtools}
\mathtoolsset{showonlyrefs}

\usepackage{cleveref}
\usepackage{arydshln}

\usepackage{algorithm}
\usepackage{algpseudocode}

\newtheorem{proposition}{Proposition}[section]

\newtheorem{lemma}[proposition]{Lemma}
\newtheorem{theorem}[proposition]{Theorem}

\theoremstyle{definition}
\newtheorem{definition}[proposition]{Definition}

\theoremstyle{remark}
\newtheorem{remark}[proposition]{Remark}

\theoremstyle{plain}

\newtheorem{assumption}[proposition]{Assumption}

\crefname{proposition}{proposition}{propositions}
\Crefname{proposition}{Proposition}{Propositions}

\crefname{lemma}{lemma}{lemmas}
\Crefname{lemma}{Lemma}{Lemmas}

\crefname{theorem}{theorem}{theorems}
\Crefname{theorem}{Theorem}{Theorems}

\crefname{definition}{definition}{definitions}
\Crefname{definition}{Definition}{Definitions}

\crefname{remark}{remark}{remarks}
\Crefname{remark}{Remark}{Remarks}

\crefname{corollary}{corollary}{corollaries}
\Crefname{corollary}{Corollary}{Corollaries}

\crefname{assumption}{assumption}{assumptions}
\Crefname{assumption}{Assumption}{Assumptions}

\AddToHook{env/lemma/begin}{\crefalias{proposition}{lemma}}
\AddToHook{env/theorem/begin}{\crefalias{proposition}{theorem}}
\AddToHook{env/definition/begin}{\crefalias{proposition}{definition}}
\AddToHook{env/remark/begin}{\crefalias{proposition}{remark}}
\AddToHook{env/corollary/begin}{\crefalias{proposition}{corollary}}
\AddToHook{env/assumption/begin}{\crefalias{proposition}{assumption}}

\def\argmin{\operatornamewithlimits{argmin}}
\newcommand{\E}{\mathbb{E}}
\newcommand{\KL}{\mathrm{KL}}

\newcommand{\Var}{\mathrm{Var}}
\def\Zset{\mathsf{Z}}
\def\Zsigma{\mathcal{Z}}
\newcommand{\MKQ}{\mathrm{P}}
\def\nset{\mathbb{N}}
\newcommand{\tmix}{\tau_{\mathrm{mix}}}
\def\eqsp{\,}
\newcommand{\tvnorm}[1]{\| #1 \|_{\operatorname{TV}}}
\def\rset{\mathbb{R}}
\def\Xset{\mathsf{X}}
\def\Xsigma{\mathcal{X}}
\def\PE{\mathsf{E}}

\newcommand{\barf}{\bar{f}}
\newcommand{\bConst}[1]{\operatorname{C}_{{\bf #1}}}

\def\rmd{\mathrm{d}}

\newcommand{\PP}{\mathbb{P}}
\def\Pens{\mathcal{P}}
\newcommand{\dsieve}{p}
\newcommand{\dsievepsi}{p_{\psi}}
\newcommand{\ERIRL}{\textsc{ER--IRL}}
\def\funcbw{\mathbf{b}}
\def\funcAw{\mathbf{A}}

\numberwithin{equation}{section}

\begin{document}

\title{Statistical analysis of Inverse Entropy-regularized
Reinforcement Learning}



\emsauthor{1}{
	\givenname{Denis}
	\surname{Belomestny}
	\mrid{}
	\orcid{}}{Belomestny}
\emsauthor{2}{
	\givenname{Alexey}
	\surname{Naumov}
	\mrid{}
	\orcid{}}{Naumov}

\emsauthor{3}{
	\givenname{Artemy}
	\surname{Rubtsov}
	\mrid{}
	\orcid{}}{Rubtsov}
    
\emsauthor{4}{
	\givenname{Sergey}
	\surname{Samsonov}
	\mrid{}
	\orcid{}}{Samsonov}

\Emsaffil{1}{
	\department{Department of Mathematics}
	\organisation{Duisburg-Essen University}
	\rorid{}
	\address{Thea-Leymann-Str. 9}
	\zip{D-45127}
	\city{Essen}
	\country{Germany}
	\affemail{denis.belomestny@uni-due.de}}
\Emsaffil{2}{
	\department{Faculty of computer sciences}
	\organisation{HSE University}
	\rorid{}
	\address{11 Pokrovsky Boulevard}
	\zip{109028}
	\city{Moscow}
	\country{Russian Federation}
	\affemail{anaumov@hse.ru}}

\Emsaffil{3}{
	\department{Faculty of computer sciences}
	\organisation{HSE University}
	\rorid{}
	\address{11 Pokrovsky Boulevard}
	\zip{109028}
	\city{Moscow}
	\country{Russian Federation}
	\affemail{asrubtsov@hse.ru}
	}
    
\Emsaffil{4}{
	\department{Faculty of computer sciences}
	\organisation{HSE University}
	\rorid{}
	\address{11 Pokrovsky Boulevard}
	\zip{109028}
	\city{Moscow}
	\country{Russian Federation}
	\affemail{svsamsonov@hse.ru}
	}



\begin{abstract}
 Inverse reinforcement learning (IRL) aims to infer the reward function that explains 
expert behavior observed through trajectories of state--action pairs. A long-standing 
difficulty in classical IRL is the non-uniqueness of the recovered reward: many reward 
functions can induce the same optimal policy, rendering the inverse problem ill-posed. 
In this paper, we develop a statistical framework for \emph{Inverse Entropy-regularized 
RL}  that resolves this ambiguity by combining entropy 
regularization with a least-squares reconstruction of the reward from the soft Bellman 
residual. This combination yields a unique and well-defined  canonical least-squares reward  \(R^\star_{\operatorname{LS}}\) consistent with the 
expert policy.
We model the expert demonstrations as a Markov chain with the invariant distribution defined by an
unknown expert policy \(\pi^\star\)
and estimate the policy by a penalized maximum-likelihood procedure over
a class of conditional distributions on the action space. We establish high-probability bounds for the
excess Kullback–Leibler divergence between the estimated policy and the expert policy, accounting for
statistical complexity through covering numbers of the policy class. These results yield non-asymptotic convergence rates for the exact
least-squares reward reconstruction, quantifying the dependence on
entropy regularization, model complexity, and sample size. We also
establish a minimax lower bound for a family of compatible rewards
with a fixed shaping function under Markov sampling. We also construct a computable trajectory-based estimator of the least-squares reward and establish finite-sample guarantees for it. Our analysis bridges behavior cloning, IRL, and modern statistical learning.
\end{abstract}

\maketitle


\tableofcontents

\section{Introduction}
\label{sec:introduction}

Inverse reinforcement learning (IRL) is concerned with the recovery of a reward 
function that explains observed expert behavior in a Markov decision process (MDP). 
Rather than prescribing optimal behavior directly, IRL assumes that the expert policy 
is optimal for some unknown reward, and aims to reconstruct this reward 
function from observed trajectories. Such formulations are particularly relevant 
in imitation learning, where the primary goal is to leverage expert demonstrations 
to reproduce or generalize expert behavior.
\par
A well-known challenge in classical IRL is that the solution is \emph{not unique}: 
many reward functions can explain the same observed expert behavior. This inherent 
ambiguity is often referred to as the ``ill-posedness'' of IRL; see e.g.,  \cite{ng2000algorithms}. Entropy regularization provides an important step towards resolving this difficulty. 
In entropy-regularized RL \cite{neu2017unifiedviewentropyregularizedmarkov}, \cite{pmlr-v97-geist19a}, the expert is assumed to optimize 
not only the expected discounted reward but also an entropy bonus, thereby producing 
smoother and more stochastic policies. This has several advantages: entropy 
regularization improves numerical stability, encourages exploration, and naturally 
leads to Gibbs (softmax) optimal policies  \cite{pmlr-v97-ahmed19a}. The corresponding ``soft'' Bellman 
equations replace the hard max operator of classical RL by the smooth log-sum-exp, 
yielding both theoretical tractability and practical robustness. In this setting, 
the optimal policy associated with a given reward is unique. However, uniqueness of the reward itself does not follow from entropy regularization 
alone. Recent identifiability results show that, even in entropy-regularized IRL, demonstrations generally determine only a feasible set of compatible rewards rather than a single ground-truth reward; see e.g., \cite{cao2021identifiability,skalse2023invariance,schlaginhaufen2024transferability,schlaginhaufen2023identifiability}. In this paper, we resolve the remaining reward ambiguity by combining
entropy regularization with a least-squares reconstruction based on the
soft Bellman residual. The resulting inverse entropy-regularized RL
procedure selects a unique canonical representative from the class of
compatible rewards. In view of the partial-identifiability results above, our least-squares step should be interpreted as selecting a canonical representative from the compatible reward class induced by the expert policy, not as removing all reward-equivalence phenomena without an additional criterion.
\par
Given expert demonstrations in the form of the Markov chain $(S_n,A_n)_{n=1}^N$ with the invariant
distribution $d_{\pi^\star}(s)\pi^\star(a \mid s)$, \ERIRL\, aims to identify a reward 
function $R$ such that the optimal entropy-regularized policy induced by $R$ 
matches the expert policy $\pi^\star$. One natural approach is to estimate the expert 
policy itself by maximum likelihood-a procedure equivalent to \emph{behavior cloning}-and 
then to back out a reward function consistent with the estimated policy. We refer to \cite{TiapkinBelomestny2024} for the analysis of behavior cloning in entropy-regularized RL. This 
yields a two-step procedure: (\emph{i}) policy estimation via penalized 
maximum-likelihood, and (\emph{ii}) reward reconstruction via least-squares formulation of the soft Bellman residual equations. Importantly, the 
regularization of the policy estimator plays a crucial role in controlling 
estimation error and ensuring identifiability of the recovered reward.
\par
\paragraph{Literature overview}
In entropy-regularized IRL, expert demonstrations are modeled as samples from a Markov chain whose transition probabilities encode both a latent reward and an entropy or KL-based regularizer.  This probabilistic structure frequently reduces reward inference to a least-squares-type statistical estimation problem, since soft Bellman relations and linearly solvable MDPs induce linear constraints linking rewards, value differences, and empirical log transition statistics. The  model-free deep IRL method from \cite{Uchibe2018} makes this structure explicit by parameterizing the log density ratio between expert and passive dynamics and estimating it via logistic regression, effectively projecting empirical Markov transition frequencies onto a parametric model of the soft-optimal dynamics. Uchibe's later work on model-based imitation learning with entropy regularization \cite{Uchibe2022} further develops this principle by jointly regularizing both the model and the policy, thereby ensuring that the induced controlled Markov chain satisfies smoothness and stability conditions that are amenable to regression-based reward estimation. These formulations highlight how entropy regularization not only facilitates robust policy inference but also supports reward estimation through linear or linearized relationships between observed transitions and latent cost parameters. A similar least-squares interpretation underlies maximum entropy deep IRL \cite{Wulfmeier2015}, where aligning empirical feature expectations with those induced by soft value iteration can be linearized as a regression step projecting empirical successor features onto the reward-induced soft occupancy measures. In continuous-control settings, the inverse optimal control approaches \cite{Levine2016} exploit Laplace approximations around locally optimal trajectories; these approximations yield linear equations in reward gradients, turning the IRL estimation step into a regularized least-squares fit to deviations of observed trajectories from locally optimal flows. The broader methodological context is summarized in the surveys \cite{ARORA2021103500},  \cite{AdamsCodyBelingSurvey}, which situate entropy-regularized IRL, density-ratio estimation, and probabilistic inverse optimal control within a unified conceptual framework. 

 A complementary line of work studies reward identifiability itself. The classical IRL algorithm of \cite{ng2000algorithms} characterizes rewards for which a policy is optimal and selects one such reward through a maximum-margin criterion. In entropy-regularized IRL, \cite{cao2021identifiability} gives a closed-form description of the feasible reward set compatible with a fixed stochastic expert policy, showing that the reward is only partially identifiable in a single environment. The invariance framework of \cite{skalse2023invariance} compares this feasible set with the ambiguity sets induced by other expert models, including optimal experts. The consequences of choosing an arbitrary compatible reward have also been analyzed: \cite{schlaginhaufen2024transferability} studies performance loss when transferring rewards from the feasible set to a new environment, while \cite{schlaginhaufen2023identifiability}  extends identifiability and generalizability questions to constrained IRL.

Although \ERIRL\, is closely related to conditional density estimation (indeed, note that an entropy-regularized policy is a conditional density of actions given states), this connection is only intermediate. Our goal is not merely to estimate the policy, but to use it as a stepping stone toward recovering the underlying reward function. This contrasts with classical conditional density estimation, where the conditional density itself is the target of inference.
The connection is nonetheless informative. Recent minimax analyses for conditional density estimation under total variation smoothness \cite{LiNeykovBalakrishnan2023} and empirical-entropy methods \cite{BilodeauFosterRoy2024} clarify the statistical difficulty of estimating action-state conditional distributions. However, \ERIRL\, introduces two fundamental deviations from the standard setting. First, reward recovery requires solving an inverse problem: the reward appears only implicitly through the soft Bellman or linearly solvable MDP relations, so error propagation from policy estimation to reward estimation must be controlled. Second, the expert demonstrations arise from a  Markov chain, rather than i.i.d. samples, which affects concentration properties. 

There is also a substantial statistical literature on learning feasible rewards. \cite{ramponi2021efficient} and \cite{metelli2021provably} study sample-efficient recovery of transferable rewards with a generative model, while \cite{metelli2023towards} provides corresponding minimax lower bounds for this line of algorithms. In the offline setting, \cite{zhao2024inverse} analyzes whether IRL is statistically harder than standard RL by estimating an analogue of the feasible reward set under data-coverage assumptions, and \cite{lazzati2024offline}  develops related pessimistic solution concepts and provably efficient offline algorithms. Finally, \cite{moulin2025inverse} proposes inverse Q-learning for offline imitation in $Q^{\pi}$-realizable MDPs. This last work is close in spirit to ours because it also exploits Bellman-type realizability, but its guarantees and assumptions are different: it targets offline imitation performance through an inverse-Q formulation, whereas our analysis studies non-asymptotic reward reconstruction for the entropy-regularized least-squares representative.

Despite these advances, the available theory does not directly cover the setting considered here. Existing identifiability results mainly characterize the whole feasible reward set, while existing statistical guarantees typically do not analyze the two-step pipeline based on estimating the expert policy from Markov-chain demonstrations and then reconstructing the least-squares soft-Bellman reward. Our contribution is therefore complementary: we provide finite-sample, high-probability and minimax guarantees for this specific entropy-regularized least-squares reconstruction framework, clarifying how Markov dependence, policy-estimation complexity, entropy smoothing and reward-identification interact.

\paragraph{Main contributions.}
Our results can be summarized as follows.
\begin{enumerate}[label=\textup{(\roman*)}]
    \item We formulate a canonical least-squares reward
    \(R_{\operatorname{LS}}^{\star}\) in the stationary expert geometry
    \(L_2(\rho^\star)\). This formulation makes the target of estimation
    explicit despite shaping non-identifiability.

    \item For a general policy class controlled by its uniform covering
    numbers, we establish a high-probability oracle inequality for the
    expected conditional KL loss of penalized maximum likelihood and
    transfer it to an \(L_2(\rho^\star)\) error bound for the exact
    least-squares reconstruction \(R_{\operatorname{LS}}^{\star}\).

    \item For H\"older-smooth soft \(Q\)-functions, we construct sparse
ReLU policy classes and derive explicit nonparametric upper bounds.
We also establish an information-theoretic lower bound under Markov
sampling for a family of compatible rewards with a fixed shaping
function.

    \item We construct a computable estimator of
    \(R_{\operatorname{LS}}^\star\) using a Galerkin approximation of the
    transition kernel and a linear two-timescale stochastic-approximation
    scheme. We prove a finite-sample bound for this estimator, separating
    policy-estimation, Galerkin, and
    stochastic-approximation errors.
\end{enumerate}

\paragraph{Paper organization.}
Section~\ref{sec:setup} introduces the entropy-regularized discounted
control problem and recalls the soft Bellman optimality equations and
the associated Gibbs policy.
Section~\ref{sec:irl} formulates the inverse problem.
Proposition~\ref{prop:reward shaping lemma} characterizes the rewards
compatible with the expert policy, while
Definition~\ref{def:canonical-LS-reward} selects a canonical
least-squares representative in the stationary geometry
\(L_2(\rho^\star)\).
The same section introduces the penalized maximum-likelihood estimator
of the expert policy and the population plug-in reward reconstruction
in Algorithm~\ref{alg:er-irl}.

Section~\ref{sec:convergence} develops the statistical theory for the
exact least-squares reconstruction.
Theorem~\ref{thm:hp-reward} establishes a high-probability oracle
inequality for the conditional Kullback--Leibler loss under Markov
sampling and transfers it to the least-squares reward error.
Theorem~\ref{thm:holder-relu-rate} specializes this result to
H\"older-smooth soft \(Q\)-functions and sparse ReLU policy classes.
Theorem~\ref{thm:pointwise-lower-bound} provides an information-theoretic
\(L_2\)-lower bound under Markov sampling.

Section~\ref{sec:estimated-bellman} treats the case in which the
transition operator and the stationary expert distribution are not
available for exact computation.
Lemma~\ref{lem:Galerkin-approximation-rate} controls the deterministic
Galerkin approximation error, and
Lemma~\ref{lem:theta_equation} reduces the projected least-squares
problem to finite-dimensional normal equations.
The plug-in conditional-centering error is controlled in
Proposition~\ref{prop:plugin-centering-error}, while
Algorithm~\ref{alg:ttsa} computes the resulting finite-dimensional
solution by linear two-timescale stochastic approximation.
Theorem~\ref{thm:estimated-Bellman-operator} combines the policy,
Galerkin, plug-in-centering, and stochastic-approximation errors.

The appendices contain the proofs of the compatible-reward
characterization, the oracle inequality, the H\"older--ReLU
specialization, the lower bound, and the Galerkin and two-timescale
stochastic-approximation results, together with additional
interpretations of the least-squares formulation.

\paragraph{Notation}

Let $(\Zset,\Zsigma)$ be a measurable space and $\Pens(\Zset)$ be the set of all probability measures on this space. For $p \in \Pens(\Zset)$ we denote by $\E_p$ the expectation w.r.t. $p$. For a random variable $\xi: \Zset \to \rset$, the notation $\xi \sim p$ means $\operatorname{Law}(\xi) = p$. We also write $\E_{p}$ instead of $\E_{\xi \sim p}$. For any $p \in \Pens(\Zset)$ and $f: \Zset \to \rset$, $p f = \E_p[f]$. For any $p, q \in \Pens(\Zset)$ the Kullback-Leibler divergence $\KL(p \mid q)$ is given by
\[
\KL(p \mid q) = \begin{cases}
\E_{p}[\log \frac{\rmd p}{\rmd q}], & p \ll q \\
+ \infty, & \text{otherwise}
\end{cases} 
\]
For a finite set $\Zset$ we denote the Shannon entropy
\(\mathcal{H}(p) \;=\; -\sum_{z \in \Zset} p(z)\log p(z)\). For \(\mathsf D \subset \rset^d\) and \(\beta>0\), let
\(\mathcal C^\beta(\mathsf D)\) denote the Banach space of
\(\beta\)-H\"older functions equipped with the norm
\begin{align}
\label{eq:holder_norm}
    \|f\|_{\mathcal C^\beta(\mathsf D)} =\sum_{|\alpha| \le \lceil \beta\rceil -1} \|\partial^\alpha f\|_\infty 
+ \sum_{ |\alpha| = \lceil\beta\rceil -1} \sup_{x \neq y \in \mathsf D} \frac{|\partial^\alpha f(x) - \partial^\alpha f(y)|}{|x - y|^{\beta - \lceil\beta\rceil + 1}_{\infty}}\eqsp,
\end{align}
where we use multi-index notation, that is $\partial^\alpha = \partial_1^{\alpha_1} \ldots \partial_d^{\alpha_d}$ with $\alpha = (\alpha_1, \ldots, \alpha_d) \in \mathbb{N}_0^d$ and $|\alpha| = \sum_{\ell=1}^d \alpha_\ell$.
We also define the ball of $\beta$-H\"older functions with radius $K$ by
\begin{align}
\mathcal C^\beta(\mathsf D, K) = \left \{f\in \mathcal C^\beta(\mathsf D): \|f\|_{\mathcal C^\beta(\mathsf D)} \le K
\right\}\eqsp.
\end{align}

\section{Entropy-regularized reinforcement learning}
\label{sec:setup}
We recall the entropy-regularized discounted control problem and fix the notation used throughout the paper. Standard background can be found, for example, in 
\cite{foster2023foundations}.
We consider an infinite-horizon discounted MDP of the form 
\[
\mathsf{M} = (\mathsf{S}, \mathsf{A}, \MKQ, R, \gamma),
\]
where $\mathsf{S}$ is the (measurable) state space, typically a Borel subset of    
$\mathbb{R}^d$, 
 $\mathsf{A}=\{1,2,\ldots,|\mathsf{A}|\}$ is a finite action set,
$\MKQ(ds' \mid s,a)=p(s' \mid s,a)\,ds'$ is a Markov kernel on $\mathsf{S}$ (the transition law),
$R:\mathsf{S}\times\mathsf{A}\to\mathbb{R}$ is a measurable reward function, 
and $\gamma\in(0,1)$ is the discount factor. 
By a Markov kernel we mean: for each $(s,a)$, $\MKQ(\cdot\mid s,a)$ is a probability measure on $(\mathsf{S},\mathcal{B}(\mathsf{S}))$ with Lebesgue density $p(\cdot\mid s,a)$, and for each measurable $B\subseteq\mathsf{S}$ the map $(s,a)\mapsto \MKQ(B\mid s,a)$ is measurable.
A stationary Markov \emph{policy} is a Markov kernel $\pi:\mathsf{S}\to\Delta(\mathsf{A})$,
\[
s \longmapsto \pi(\cdot\mid s)=\big(\pi(a\mid s)\big)_{a\in\mathsf{A}},
\qquad 
\pi(a\mid s)\ge 0,\ \ \sum_{a\in\mathsf{A}}\pi(a\mid s)=1.
\]
We work throughout with stationary policies; in the discounted setting this is without loss of generality for optimal control.
Given an initial state $S_0\sim \mu_0$, a policy $\pi$, and the transition kernel $\MKQ$,
the process evolves as $A_t\sim \pi(\cdot\mid S_t)$ and $S_{t+1}\sim \MKQ(\cdot\mid S_t,A_t)$.
The (unregularized) discounted return is
\[
G^\pi \;=\; \sum_{t=0}^{\infty}\gamma^t\, R(S_t,A_t),
\qquad 
V^\pi(s)\;=\; \mathbb{E}\!\left[\,G^\pi\mid S_0=s\,\right].
\]
The usual unregularized performance criterion is $V^\pi(s)$. In contrast, in entropy-regularized reinforcement learning (see \cite{neu2017unifiedviewentropyregularizedmarkov}, \cite{pmlr-v97-geist19a},\cite{Mei2019}) to encourage exploration and produce smoother control laws, we add a Shannon-entropy bonus with temperature $\lambda>0$:
\[
J_\lambda(\pi) \;=\; 
\mathbb{E}\,\Bigg[\sum_{t=0}^\infty \gamma^t \Big(R(S_t,A_t) + \lambda\,\mathcal{H}(\pi(\cdot\mid S_t))\Big) \ \Big | \ S_0\sim\mu_0\Bigg]
\]
with \(\mathcal{H}(\pi(\cdot\mid s)) \;:=\; -\sum_{a\in\mathsf{A}}\pi(a\mid s)\log\pi(a\mid s).\)
Define the \emph{entropy-regularized value function} and $Q$-function:
\[
V_\lambda^\pi(s) \;=\; \E\Bigg[\sum_{t=0}^\infty \gamma^t \Big(R(S_t,A_t) + \lambda\,\mathcal{H}(\pi(\cdot\mid S_t))\Big)\ \Big|\ S_0=s\Bigg],
\]
and
\[
Q_\lambda^\pi(s,a) \;=\; R(s,a)+\gamma \MKQ V_\lambda^\pi(s, a),
\]
where for $f: S \to \rset$ we denote $\MKQ f(s,a) := \E_{S' \sim \MKQ(\cdot | s,a)}[f(S')]$. 
The goal is to maximize $J_\lambda(\pi)$ (equivalently $V_\lambda^\pi(s)$ pointwise in $s$) over policies. Following Theorems 1 and 2 in \cite{pmlr-v70-haarnoja17a},  we can show that
the optimal value function $V^\star\equiv V^\star_\lambda$ solves the entropy-regularized Bellman optimality equation
\begin{equation}
\label{eq:bellman}
V^\star(s) \;=\; \max_{\pi(\cdot|s)\in\Delta(\mathsf{A})}  \left\{\sum_{a'} \pi(a'|s)  Q^\star(s,a') + \lambda\, \mathcal{H}(\pi(\cdot|s))\right\}\eqsp,
\end{equation}
where 
\begin{equation}
\label{eq:QR}
Q^\star(s,a) \;=\; R(s,a) + \gamma \MKQ  V^\star(s,a).
\end{equation}
Note that \eqref{eq:bellman} is a concave optimization problem over the simplex $\Delta(\mathsf{A})$ with a closed-form solution. 
By the variational identity for the log-sum-exp (Gibbs variational principle),
we obtain the \emph{soft Bellman optimality equation}
\begin{equation}
\label{eq:soft-bellman}
V^\star(s) \;=\; \lambda \log \sum_{a\in\mathsf{A}} \exp\!\Big(Q^\star(s,a)/\lambda\Big)\eqsp.
\end{equation}
Furthermore, the maximizer is the Gibbs/softmax distribution
\begin{equation}
\label{eq:softmax}
\pi^\star(a\mid s) \;=\; 
\frac{\exp\!\big(Q^\star(s,a)/\lambda\big)}{\sum_{a'\in\mathsf{A}}\exp\!\big(Q^\star(s,a')/\lambda\big)}.
\end{equation}
Notice that as $\lambda \to 0$ we recover the well-known Bellman equations (see e.g., \cite{sutton2018reinforcement}) since the log-sum-exp converges to the maximum and \eqref{eq:softmax} approximates a uniform distribution over $\arg \max$. 
It is often convenient to write a ``soft advantage''
\begin{equation}
\label{eq: soft advantage}
A^\star(s,a) \;=\; Q^\star(s,a) - V^\star(s),
\end{equation}
so that \eqref{eq:softmax} becomes
\(
\pi^\star(a\mid s) \;=\; \exp\!\Big(\tfrac{1}{\lambda}A^\star(s,a)\Big),
\)
since
\(\sum_{a'}\exp\!\big(Q^\star(s,a')/\lambda\big) = \exp\!\big(V^\star(s)/\lambda\big)\).


\section{Inverse Entropy-regularized Reinforcement Learning}
\label{sec:irl}

In the entropy-regularized setting, we assume that the observed behavior is optimal for some unknown reward function 
\(
R^\star: \mathsf{S} \times \mathsf{A} \;\rightarrow\; \mathbb{R}.
\)
Let 
\(
\mathsf{M}_0 = (\mathsf{S}, \mathsf{A}, \gamma, \lambda)
\)
denote the known components of the MDP $\mathrm{M}$: the state space $\mathsf{S}$, action set $\mathsf{A}$, discount factor $\gamma\in(0,1)$, and entropy weight (temperature) $\lambda>0$. Suppose we observe a Markov chain  
\(Z_n = 
(S_n,A_n)
\) on the space \(\Zset = \mathsf{S}\times\mathsf A\).
Let \(d_{\pi^\star}\) denote the invariant state distribution under
\(\pi^\star\), that is,
\[
    d_{\pi^\star}(B)
    =
    \int_{\mathsf S}
    \sum_{a\in\mathsf A}
    \pi^\star(a\mid s)\MKQ(B\mid s,a)\,
     d_{\pi^\star}(ds),
    \qquad
    B\in\mathcal B(\mathsf S).
\]
The invariant state--action distribution is
\[
    \rho^\star(ds,da)
    :=
    d_{\pi^\star}(ds)\pi^\star(da\mid s)\eqsp,
\]
where $\pi^\star$  is the expert's conditional action distribution induced by an unknown reward $R^\star$. Let $\MKQ^{\pi^\star}$ be the transition kernel of the joint chain $Z_n$,
 that is,
\[
    \MKQ^{\pi^\star}((s,a),ds'\,da')
    =
    \MKQ(ds'\mid s,a)\,\pi^\star(da'\mid s').
\] For a reward function $
R: \mathsf{S}\times\mathsf{A} \to \mathbb{R}$ the optimal soft policy induced by $R$ is given by
\[
\pi_R^\star(a\mid s)
\;=\;
\frac{\exp\!\left(Q_R^\star(s,a)/\lambda\right)}
{\sum_{a'\in\mathsf{A}} \exp\!\left(Q_R^\star(s,a')/\lambda\right)},
\]
where $Q_R^\star(s,a) \;=\; R(s,a) + \gamma \MKQ V_R^\star(s,a)$ and $V_R^\star$ is defined in \eqref{eq:soft-bellman} .
Our goal is to recover a reward function \(R\) such that the corresponding optimal policy
\(\pi^\star_R\) matches the expert action distribution 
\begin{align}
\label{eq:compatible}
\pi^\star_R(da\mid s) \;=\; \pi^\star(da\mid s)\eqsp
\end{align}
for \(d_{\pi^\star}\)-almost every \(s\).
We refer to reward functions \(R\in L_2(\rho^\star)\) satisfying
\eqref{eq:compatible} as \emph{compatible rewards}. A natural approach to
recovering such a reward is to maximize the log-likelihood of the expert
samples under the candidate optimal policy:
\[
\hat{R}
\;\in\;
\arg\max_{R \in \mathcal{R}}
\;
\frac{1}{N} \sum_{n=1}^N 
\log \pi^\star_R(A_n \mid S_n)
\;-\;
\Omega(R),
\]
where $\Omega(R)$ is a regularizer to ensure identifiability or to encourage smoothness or sparsity of the learned reward and $\mathcal R$ is some reward class. In this paper, we use a different approach based on a more explicit relation between \(\pi^\star_R\) and \(R\) in the Entropy-regularized RL.

\subsection{Compatible rewards and the canonical least-squares representative}
\label{subsec:canonical-LS}

Entropy regularization makes the optimal policy unique for a fixed reward, but the converse map remains non-identifiable. The following statement holds; see also \cite{cao2021identifiability}.  

\begin{proposition}
\label{prop:reward shaping lemma}
The class of rewards compatible with \(\pi^\star\) consists precisely of
the functions
\begin{equation}
\label{eq:R_f}
R_f(s,a) \;:=\; f(s) + \lambda \log \pi^\star(a \mid s) 
- \gamma \MKQ f(s, a)\eqsp, \qquad f\in L_2(d_{\pi^\star})\eqsp.
\end{equation}
In particular, for every \(f\in L_2(d_{\pi^\star})\), the reward \(R_f\)
induces \(\pi^\star\) as its optimal entropy-regularized policy.
\end{proposition}
\begin{proof}
    The proof is given in Appendix~\ref{app:compatible}.
\end{proof}
Set
\(
    r^\star(s,a):=\lambda\log\pi^\star(a\mid s).
\)
To characterize the class of rewards compatible with $\pi^\star$, consider the Hilbert spaces
\begin{equation}
\label{eq:HG-spaces}
    \mathbb H:=L_2(\mathsf S,d_{\pi^\star}),
    \qquad
    \mathbb G:=L_2(\mathsf S\times\mathsf A,\rho^\star),
\end{equation}
together with the canonical embedding
\[
    \mathcal I:\mathbb H\to\mathbb G,
    \qquad
    (\mathcal I f)(s,a):=f(s).
\]
The operator $\mathcal I$ is an isometry, with adjoint
\begin{equation}
\label{eq:I-adjoint}
    (\mathcal I^\ast g)(s)
    =
    \sum_{a\in\mathsf A}\pi^\star(a\mid s)g(s,a),
    \qquad
    \mathcal I^\ast\mathcal I=\operatorname{Id}_{\mathbb H}.
\end{equation}
Further, define \(\mathcal M:=\mathcal I-\gamma\MKQ\eqsp.\)
Then, by \Cref{prop:reward shaping lemma}, every reward compatible with $\pi^\star$ has the form
\(    r^\star+\mathcal M f.
\)
By Jensen's inequality and stationarity of \(d_{\pi^{\star}}\),
\begin{align}
\label{eq:P-is-contraction}
    \|\MKQ f\|_{\mathbb G}^{2}
    &\leq
    \int_{\mathsf S\times\mathsf A}
    \int_{\mathsf S}f(s')^2 \MKQ (ds'\mid s,a)\rho^\star(ds,da)
    =
    \int_{\mathsf S}f(s')^2d_{\pi^{\star}}(ds')
    =
    \|f\|_{\mathbb H}^{2}.
\end{align}
Consequently,
\begin{equation}
\label{eq:M-lower-bound}
    \|\mathcal M f\|_{\mathbb G}
    \geq
    (1-\gamma)\|f\|_{\mathbb H}.
\end{equation}
In particular, $\mathcal M$ is injective and $\operatorname{Ran}(\mathcal M)$ is closed. We use the stationary expert law to fix the geometry of the inverse problem.

\begin{definition}[Canonical least-squares reward]
\label{def:canonical-LS-reward}
Define
\begin{equation}
\label{eq:population-LS-reward}
    f_{\mathrm{LS}}^\star
    :=
    \argmin_{f\in\mathbb H}
    \|r^\star+\mathcal M f\|_{\mathbb G}^2,
    \qquad
    R_{\mathrm{LS}}^\star
    :=
    r^\star+\mathcal M f_{\mathrm{LS}}^\star.
\end{equation}
\end{definition}
The minimizer is unique by \eqref{eq:M-lower-bound}; equivalently, \(R_{\mathrm{LS}}^\star
    =
    (\operatorname{Id}_{\mathbb G}-\Pi_{\mathcal M})r^\star,\)
where $\Pi_{\mathcal M}$ is the orthogonal projector onto $\operatorname{Ran}(\mathcal M)$ in $\mathbb G$. This is the precise sense in which the least-squares criterion selects a canonical representative of the shaping class.

\subsection{Policy estimation and plug-in reward reconstruction}
\label{sec:algorithm}
Consider a  class of candidate policies $\mathcal{F}$ 
\[
\mathcal{F} \;\subseteq\; 
\Big\{
\pi: \mathsf{S} \to \Delta(\mathsf{A})
\;\big|\;
\pi(\cdot\mid s) \text{ measurable in } s
\Big\}.
\]
Given observations from the uniformly ergodic Markov chain
\((S_n,A_n)_{n=1}^N\), define the penalized maximum-likelihood estimator by
\begin{equation}
\label{hat pi def}
    \widehat\pi
    \in
    \argmin_{\pi\in\mathcal F}
    \left\{
        \widehat L_N(\pi)+\mathcal R(\pi)
    \right\},
    \qquad
    \widehat L_N(\pi)
    :=
    -\frac1N\sum_{n=1}^N\log\pi(A_n\mid S_n).
\end{equation}
Here $\mathcal R(\pi)\geq 0$ is a regularization term (e.g., a norm penalty, entropy regularization, or other complexity-control term).
The population counterpart of the empirical risk is
\begin{multline}
\label{eq: L def}
L(\pi)
= \E_{(S,A)\sim \rho^\star}[-\log \pi(A\mid S)]
\\
= \E_{S \sim d_{\pi^\star}}\!\left[
\mathrm{KL}(\pi^\star(\cdot\mid S)\mid \pi(\cdot\mid S))
\right]
+ \E_{S \sim d_{\pi^\star}}\!\left[
\mathcal{H}(\pi^\star(\cdot\mid S))
\right],  
\end{multline}
where $\mathcal{H}$ is the Shannon entropy.
Since the entropy term does not depend on $\pi$, minimizing $L(\pi)$ is equivalent to minimizing expected conditional KL divergence
\begin{align}
\label{eq:kl_pi}
\mathcal{K}(\pi) \; := \;\E_{S \sim d_{\pi^\star}}\!\left[
\mathrm{KL}(\pi^\star(\cdot\mid S) \mid \pi(\cdot\mid S))
\right].
\end{align}
Thus, in the infinite-data limit, the minimizer of $L(\pi)$ is $\pi^\star$ itself.
This approach is known as \emph{behavior cloning} in imitation learning.
Set
\begin{equation}
\label{eq:rhat-def}
    \widehat r(s,a):=\lambda\log\widehat\pi(a\mid s)
\end{equation}
and define the population plug-in reward by
\begin{equation}
\label{eq:Rhat-known-P}
    \widehat R
    :=
    (\operatorname{Id}_{\mathbb G}-\Pi_{\mathcal M})\widehat r.
\end{equation}
Finally, we can combine behavior cloning with the least-squares reward fitting to select an appropriate reward function. Both procedures are combined in Algorithm~\ref{alg:er-irl}.
\begin{algorithm}[H]
\caption{Population \ERIRL\ reconstruction}
\label{alg:er-irl}
\begin{algorithmic}[1]
\Require Expert trajectory $\mathcal D = (S_n,A_n)_{n=1}^N$, policy class $\mathcal F$, penalty $\mathcal R$, and $\mathrm M_0$.
\State Compute $\widehat\pi$ from \eqref{hat pi def}.
\State Set $\widehat r=\lambda\log\widehat\pi$.
\State
\label{step:reward_fitting}
Construct $\widehat R$ according to \eqref{eq:Rhat-known-P}.
\State \Return $\widehat\pi$ and $\widehat R$.
\end{algorithmic}
\end{algorithm}

Note that \Cref{alg:er-irl} defines the \ERIRL{} procedure at a general level and does not prescribe how step~\ref{step:reward_fitting} should be implemented. In practice, the difficulty of this step stems from the fact that both the transition kernel \(\MKQ\) and the distribution \(\rho^\star\) are unknown. This coupled uncertainty naturally motivates a two-timescale stochastic approximation scheme (see \Cref{alg:ttsa}), which replaces the exact computation in step~\ref{step:reward_fitting}.
We refer the reader to Section~\ref{sec:estimated-bellman} for a detailed discussion.

\section{Statistical guarantees for an exact least-squares solution}
\label{sec:convergence}
\subsection{A general upper bound}
\label{subsec:general-oracle}
We make the following assumptions.

\begin{assumption}
    \label{assum: MDP}
    \(\mathsf{M}_0 = (\mathsf{S}, \mathsf{A}, \gamma, \lambda)\) is given. The state space \(\mathsf S\) is a  measurable subset of $\rset^d$ and $\mathsf{A}$ is finite. 
\end{assumption}

\begin{assumption}[Uniform geometric ergodicity]
\label{assum:UGE}
Let \(\rho_0\) be the initial distribution of the joint chain
\(Z_n=(S_n,A_n)\). Assume that \(\rho_0\ll\rho^\star\) and define
\begin{equation}
\label{eq:initial-density-ratio}
    \mathfrak N_0
    :=
    \int_{\mathsf S\times\mathsf A}
    \left(
        \frac{d\rho_0}{d\rho^\star}
    \right)^2
    d\rho^\star
    <\infty.
\end{equation}
Moreover, there exists $\tmix\in\mathbb N$ such that, for every $n\geq1$ and $(s,a)\in\mathsf S\times\mathsf A$,
\begin{equation}
\label{eq:tv_ergodicity-as}
    \Big\|\big( \MKQ^{\pi^\star}\big)^n((s,a),\cdot)-\rho^\star\Big\|_{\operatorname{TV}}
    \leq
    (1/4)^{\lfloor n/\tmix\rfloor}.
\end{equation}
\end{assumption}
The parameter $\tmix$ in \Cref{assum:UGE} is referred to as a \emph{mixing time}, see, e.g., \cite{paulin2015concentration}, and controls the rate of convergence of the iterates $\big(\MKQ^{\pi^\star}\big)^n$ to $\rho^\star$ as $n$ increases. Note that \Cref{assum:UGE} is a standard assumption in the theory of Markov chains with a compact state space.

\begin{assumption}
    \label{assum: F}
Let \(\mathcal F\subset \ell_\infty(\mathsf S\times\mathsf A)\) be a non-empty class of policies. Assume that \(\mathcal F\) is
closed in \(\|\cdot\|_\infty\), that
\(\mathcal R:\mathcal F\to[0,\infty]\) is lower semicontinuous, and that
there exists \(\alpha\in(0,1/|\mathsf A|]\) such that
\(
    \pi(a\mid s)\geq\alpha 
\) for all \(\pi\in\mathcal F\)\ and 
    \((s,a)\in\mathsf S\times\mathsf A.\)
Assume, moreover, that there exist \(D\geq1\) and \(\Lambda>0\) such that,
for every \(\varepsilon\in(0,\Lambda]\),
\begin{equation}
\label{eq:entropy-bound}
    \log\mathcal N
    \bigl(\varepsilon,\mathcal F,\|\cdot\|_\infty\bigr)
    \leq
    D\log\left(\frac{\Lambda}{\varepsilon}\right).
\end{equation}
Here \(\mathcal N\big(\varepsilon,\,\mathcal F,\,\|\cdot\|_\infty\big)\) denotes the $\varepsilon$-covering number of $\mathcal F$ with respect to
$\|\cdot\|_\infty$.
\end{assumption}
\begin{remark}
\label{rem:compact}
    The entropy bound implies that \(\mathcal F\) is totally bounded. Since
\(\mathcal F\) is closed in the complete space
\(\ell_\infty(\mathsf S\times\mathsf A)\), it is compact. The empirical
negative log-likelihood is continuous on \(\mathcal F\) by the uniform
floor, while \(\mathcal R\) is lower semicontinuous. Hence the criterion
in \eqref{hat pi def} attains its minimum. The same argument shows that
the infimum in \eqref{eq:Delta_F} is attained.
\end{remark}
 \Cref{assum: F} is a standard requirement on the covering number of a parametric class. For example, below we show that the conditions will be satisfied if $\mathcal F$ is the class of feed-forward neural networks and $Q^\star$-functions belong to the ball of Hölder functions and their range of values is bounded. See \Cref{thm:holder-relu-rate} for a concrete construction of such a policy class.

Define
\begin{equation}
\label{eq:Delta_F}
\Delta(\mathcal F)
=\inf_{\pi\in\mathcal F}\Big\{ \mathcal{K}(\pi)+\mathcal R(\pi)\Big\}.
\end{equation}

For \(\delta\in(0,1)\), define
\begin{align}
\label{eq:first-stage-rate}
        \mathfrak C_N(\delta)
    &:=
    \left(D\log\left(
        \frac{1+\Lambda N/\tau_{\mathrm{mix}}}{\alpha}
    \right)
    +
    \log\left(\frac{\sqrt{2\mathfrak N_0}}{\delta}\right)\right)\left(\log\frac1\alpha+2\right)\eqsp.
\end{align} 
\begin{theorem}
\label{thm:hp-reward}
Let \Cref{assum: MDP,assum: F,assum:UGE} hold and $N\geq2\tmix$. For $\delta\in(0,1)$, with probability at least $1-\delta$, the following bounds hold:
\begin{equation}
\label{eq:KL-eps}
\begin{aligned}
    &\mathcal{K}(\widehat\pi)
   \lesssim
    \left(\sqrt{\Delta(\mathcal F)}
    +
    \sqrt{
        \frac{\mathfrak C_N(\delta)\tmix}{N}
    }
    \right)^2\eqsp, 
\end{aligned}
\end{equation}
Moreover, 
\begin{equation}
\label{eq:hp-reward}
    \|\widehat R-R_{\mathrm{LS}}^\star\|_{\rho^\star}^2
    \lesssim
    \lambda^2\left(\log\frac1\alpha+2\right)
    \left[
        \Delta(\mathcal F)
        +
        \frac{\mathfrak C_N(\delta)\tmix}{N}
    \right].
\end{equation}
\end{theorem}
\begin{proof}
    The proof is given in Appendix~\ref{subsec:policy-oracle-proof}.
\end{proof}
The first inequality is an oracle inequality for conditional maximum likelihood under Markov sampling. The second is a direct stability consequence of the orthogonal projection defining the canonical reward: policy error is not amplified by the Bellman projection in the $L_2(\rho^\star)$ geometry, apart from the logarithmic moment factor needed to pass from KL loss to squared log-policy loss.
We next instantiate this oracle inequality with sparse ReLU policy classes.

\subsection{H\"older-smooth logits and sparse ReLU policies}
\label{subsec:relu_policies}
In this section we assume that  $\mathsf{S}=[0,1]^d$ and $\mathcal R = 0$. We impose an additional assumption on $Q^\star$. 
\begin{assumption}
\label{assum:Holder}
$Q^\star(\cdot,a) \in \mathcal C^\beta([0,1]^d, K)$ for every $a \in \mathsf{A}$ and some $\beta > 0$ and $K > 0$.
\end{assumption}
Following \cite{MR4134774}, we show that this assumption guarantees that one may construct a feed-forward neural network that approximates $Q^\star$ with given precision. 
\begin{assumption}
\label{assum: range}
There exists $\Gamma\ge 0$ such that for all $s\in\mathsf S$,
\begin{equation}
\label{eq:spanqstar}
\operatorname{span}_a Q^\star(s,\cdot)
:= \max_{a} Q^\star(s,a) - \min_{a} Q^\star(s,a) \;\le\; \Gamma \, .
\end{equation}
\end{assumption}
In particular, we note that this assumption implies that for all $(s,a) \in \mathsf{S} \times \mathsf{A}$,
\[
\pi^\star(a\mid s)\ \ge\ \frac{1}{\,1+(|\mathsf A|-1)\,e^{\Gamma/\lambda}\,}.
\]

\begin{theorem}
\label{thm:holder-relu-rate}
 Let \Cref{assum: MDP,assum:UGE,assum:Holder,assum: range} hold and $\mathsf{S}=[0,1]^d$. Then, for every sufficiently large \(N\geq2\tmix\), there exists a policy
class \(\mathcal F_N\) satisfying \Cref{assum: F}. Let
\(\widehat\pi\) be the maximum-likelihood estimator over
\(\mathcal F_N\). For every $\delta\in(0,1)$, there exists a constant $c>0$ independent of $N$ such that, with probability at least $1-\delta$,  
\[
\mathcal{K}(\widehat\pi) \lesssim  K^{\frac{2d}{d+2\beta}}\lambda^{-\frac{2d}{d+2\beta}}\,
\Big(\frac{ |\mathsf{A}| \tmix}{N}\Big)^{\frac{2\beta}{d+2\beta}} \log^c N  + \frac{\log({\sqrt{2\mathfrak N_0}}/{\delta}) \tmix}{N},
\]
and 
\[
\|\widehat R - R_{\mathrm{LS}}^\star\|_{L_2(\rho^\star)}^2 \lesssim 
K^{\frac{2d}{d+2\beta}} \Big(\frac{ \lambda^2|\mathsf{A}|\tmix}{N}\Big)^{\frac{2\beta}{d+2\beta}} \log^c N  + \frac{\lambda^2 \log({\sqrt{2\mathfrak N_0}}/{\delta}) \tmix}{N} \, .
\]
Here $\lesssim$ stands for inequality up to absolute and problem-specific constants (such as $d, \beta, \Gamma$, etc.), but not $N, K$ and $\lambda$.  
\end{theorem}
\begin{proof}
    The proof is given in Appendix~\ref{app:proof-holder-relu}.  
\end{proof}

\subsection{Minimax lower bound under Markov sampling}
\label{subsec:lower-bound-main}

Throughout this subsection, let \(\mathsf S=[0,1]^d\) and
\(A:=|\mathsf A|\geq2\).To remove the statewise additive ambiguity of the
softmax parametrization, introduce the centered H\"older class
\begin{equation}
\label{eq:centered-logit-class}
    \mathcal G^\beta(K)
    :=
    \left\{
        g:\mathsf S\times\mathsf A\to\mathbb R:
        \max_{a\in\mathsf A}
        \|g(\cdot,a)\|_{\mathcal C^\beta(\mathsf S)}
        \leq K,\quad
        \sum_{a\in\mathsf A}g(s,a)=0
        \text{ for every }s
    \right\}.
\end{equation}
For \(g\in\mathcal G^\beta(K)\), define
\begin{equation}
\label{eq:lower-bound-policy-reward}
    \pi_g(a\mid s)
    :=
    \frac{\exp(g(s,a)/\lambda)}
         {\sum_{b\in\mathsf A}\exp(g(s,b)/\lambda)},
    \qquad
    r_g(s,a)
    :=
    \lambda\log\pi_g(a\mid s).
\end{equation}
Fix a bounded measurable shaping function
\(f_0:\mathsf S\to\mathbb R\), and set
\begin{equation}
\label{eq:lower-bound-compatible-reward}
    R_{g,f_0}
    :=
    r_g+\mathcal M f_0.
\end{equation}
By \Cref{prop:reward shaping lemma}, \(R_{g,f_0}\) induces the policy
\(\pi_g\). For every \(g\), let \(\mathbb P_g^{(N)}\) denote the law of the
trajectory \(\mathcal D_N
    =
    (S_0,A_0,\ldots,A_{N-1},S_N)\)
generated according to
\(
    S_0\sim\mu_0\)
    ,
    \(
    A_t\sim\pi_g(\cdot\mid S_t)\) and 
    \(S_{t+1}\sim\MKQ(\cdot\mid S_t,A_t),
\)
where the initial state distribution \(\mu_0\) does not depend on \(g\).

\begin{assumption}
\label{assum:lower-bound-dynamics}
The transition kernel admits a Lebesgue density
\(p(s'\mid s,a)\) satisfying
\begin{equation}
\label{eq:two-sided-transition-density}
    0<\underline p
    \leq
    p(s'\mid s,a)
    \leq
    \overline p<\infty
\end{equation}
for any \(s,s'\in[0,1]^d\) and \(a\in\mathsf A\).
Moreover, \(\mu_0\) admits a density \(q_0\) such that \(\|q_0\|_\infty\leq\overline q_0.\)
\end{assumption}

Under \eqref{eq:two-sided-transition-density}, the state kernel induced
by every policy \(\pi_g\) has a unique invariant distribution with
density \(q_g\) satisfying \( \underline p
    \leq
    q_g(s)
    \leq
    \overline p.\)
We write
\[
    \rho_g(ds,da)
    :=
    q_g(s)\pi_g(a\mid s)\,ds.
\]

\begin{theorem}
\label{thm:pointwise-lower-bound}
Suppose that \Cref{assum:lower-bound-dynamics} holds and \(\mathsf S=[0,1]^d\). There exists a 
constant \(c >0\), depending only on \(d,
    \beta,
    \underline p,
    \overline p,
    \overline q_0\),
such that, for all \(N\ge N_0\),
one has
\begin{equation}
\label{eq:markovian-L2-lower-bound}
    \inf_{\widehat R_N}
    \sup_{g\in\mathcal G^\beta(K)}
    \mathbb E_g
    \left[
        \|
            \widehat R_N-R_{g,f_0}
        \|_{\rho_g}^2
    \right]
    \geq
    c
    K^{\frac{2d}{2\beta+d}}
    \left(
        \frac{\lambda^2 A}{N}
    \right)^{\frac{2\beta}{2\beta+d}}.
\end{equation}
The infimum is over all estimators measurable with respect to
\(\mathcal D_N\) and \(N_0\) is given in \eqref{eq:lower-bound-sample-size}.
\end{theorem}
\begin{proof}
    The proof is given in Appendix~\ref{app:proof-lower-bound}.  
\end{proof}

\section{Unknown transition operator: Galerkin approximation and
two-timescale stochastic approximation}
\label{sec:estimated-bellman}

In this section, we assume that 
 \(\mathsf S=[0,1]^d\).
\Cref{thm:hp-reward} provides a statistical guarantee for the exact least-squares estimator $\widehat R$. However, this estimator is not directly computable, since both the transition operator and the expert stationary distribution are unknown. In this section, we  develop a computable approximation to Step~\ref{step:reward_fitting} in \Cref{alg:er-irl}.

We begin with the following four-term error decomposition:
\begin{align}
\label{eq:four-term-decomposition}
    \|R_{\mathrm{LS}}^\star-\widehat R_{\dsieve}^{\mathrm{SA}}\|_{\rho^\star}
    &\leq
    \underbrace{
        \|R_{\mathrm{LS}}^\star-\widehat R\|_{\rho^\star}
    }_{\text{policy estimation}}
    +
    \underbrace{
        \|\widehat R-\widehat R_\dsieve\|_{\rho^\star}
    }_{\text{Galerkin approximation}}
    \\
    &\quad+
    \underbrace{
        \|\widehat R_\dsieve-\widehat R_{\dsieve,\widehat\pi}\|_{\rho^\star}
    }_{\text{plug-in error}}
    +
    \underbrace{
        \|\widehat R_{\dsieve,\widehat\pi}
        -\widehat R_{\dsieve}^{\mathrm{SA}}\|_{\rho^\star}
    }_{\text{stochastic approximation}}.
\end{align}
Here, the policy estimation term involves the exact least-squares estimator
\(\widehat R\) and has already been analysed in \Cref{thm:hp-reward}.
To reduce the projection problem to a finite-dimensional one, we approximate
the kernel \(\MKQ\) using the Galerkin projection method and denote by
\(\widehat R_{\dsieve}\) the corresponding exact solution with the projected
kernel. This gives rise to the Galerkin approximation term, which is analysed
in \Cref{subsec:projection-based-method}. The plug-in error arises when the
expert policy \(\pi^\star\), which enters the coefficients of the
finite-dimensional representation of the reward, is replaced by its estimate
\(\widehat\pi\). The resulting estimator is denoted by
\(\widehat R_{\dsieve,\widehat\pi}\), and the corresponding error is analysed
in \Cref{prop:plugin-centering-error}. Finally, we approximate
\(\widehat R_{\dsieve,\widehat\pi}\) using the TTSA algorithm in \Cref{alg:ttsa} and denote
its output after \(N\) iterations by
\(\widehat R_{\dsieve}^{\mathrm{SA}}\), see \Cref{subsec:finite-dimensional-projection}. The resulting error bound is stated in
\Cref{thm:estimated-Bellman-operator}.

\subsection{Projection-based approximation and error decomposition}
\label{subsec:projection-based-method}
Our approach is based on Galerkin approximations of the transition kernel
\(\MKQ\). To quantify the resulting approximation error, we impose the
following regularity assumption on the transition densities.
\begin{assumption}
\label{assum:kernel_smoothness}
    For each \(a\in\mathsf A\), let \(\kappa_a(s,\cdot)\) be the density of \(\MKQ(\cdot\mid s,a)\) with respect to \(d_{\pi^\star}\):
    \[
        \MKQ(ds'\mid s,a)
        =
        \kappa_a(s,s')\,d_{\pi^\star}(ds').
    \]
   Moreover, for every \(a\in\mathsf{A}\), \(\kappa_a\) satisfies the following \(\beta\)-Hölder regularity condition:
    \begin{align}
    \label{eq:kernel-output-smoothness}
        L_\kappa^2
:=
\Big(
    &\max_{a\in\mathsf A}
    \int_{\mathsf S}
    \|\kappa_a(\cdot,s')\|_{\mathcal C^\beta(\mathsf S)}^2
    \,d_{\pi^\star}(ds')\Big)\vee
    \int_{\mathsf S\times\mathsf A}
    \|\kappa_a(s,\cdot)\|_{\mathcal C^\beta(\mathsf S)}^2
    \,\rho^\star(ds, da)
<\infty.
\end{align}
\end{assumption}
For a fixed \(\dsieve\geq1\), introduce the \(\dsieve\)-dimensional sieve subspace
\(\mathcal V_{\dsieve}\subset\mathbb H\) with basis
\(\Phi_{\dsieve}(s):=(\phi_1(s),\ldots,\phi_{\dsieve}(s))^\top\).
For the state--action space, define
\begin{align}
\label{eq:Psi_definition}
    \mathcal W_{\dsieve}
    :=
    \oplus_{a\in\mathsf A}\mathcal V_{\dsieve}
    \subset\mathbb G,
    \qquad
    \psi_{j,a_0}(s,a)
    :=
    \phi_j(s)\mathbf 1_{\{a=a_0\}},
\end{align}
for \(j=1,\ldots,\dsieve\) and \(a_0\in\mathsf A\). Hence,
\(\dsievepsi:= \dim(\mathcal W_{\dsieve})=\dsieve|\mathsf A|\). 

For \(\beta>0\), let
\(\mathcal C^\beta(\mathsf S)\) denote the usual Hölder space on
\(\mathsf S\) with norm \(\|\cdot\|_{\mathcal{C}^\beta(\mathsf{S})}\). 
We impose the following approximation and boundedness conditions on the sieve spaces and their bases.
\begin{assumption}
\label{assum:kernel_approximation}
    The sieve space \(\mathcal V_\dsieve\) satisfies the Jackson-type approximation property
    \begin{align}
    \label{eq:Jackson-Vk}
       \inf_{v\in\mathcal{V}_\dsieve} \|f - v\|_{\infty}
        &\leq
        C_{\mathcal V}
        \dsieve^{-\beta/d}
        \|f\|_{\mathcal C^\beta(\mathsf S)},
        \qquad f\in\mathcal C^\beta(\mathsf S),
    \end{align}
    For every \(\dsieve\), the feature vectors are linearly independent in \(L_2(d_{\pi^\star})\) and  uniformly
    bounded:
    \begin{equation}
    \label{eq:bounded-features}
        \mathfrak B_{\phi,\dsieve}
        :=
        \sup_{s\in\mathsf S}
        \|\Phi_{\dsieve}(s)\|
        <\infty,
    \end{equation}
\end{assumption}

\paragraph{Examples of spaces satisfying \Cref{assum:kernel_approximation}}
The approximation properties in \eqref{eq:Jackson-Vk} are satisfied by
many standard approximation spaces. Suppose first that
\(\mathsf S=[0,1]^d\). Let \(\mathcal S_J\) denote a resolution-\(J\)
space on \(\mathsf S\). Standard Jackson-type inequalities give
\begin{equation}
\label{eq:examples-Jackson-resolution}
    \inf_{v\in\mathcal S_J}
    \|f-v\|_{\infty}
    \leq
    C J^{-\beta}
    \|f\|_{\mathcal C^\beta(\mathsf S)}
\end{equation}
for tensor-product algebraic polynomials of coordinatewise degree at
most \(J\)
\cite[Ch.~7, Theorem~6.1]{devoreLorentz1993constructive}
and for tensor-product trigonometric polynomials with frequencies
\(\|\ell\|_\infty\leq J\)
\cite[Ch.~7, Theorem~2.1]{devoreLorentz1993constructive},
provided in the latter case that \(f\) admits a periodic
\(\mathcal C^\beta\)-extension. The same estimate holds for tensor-product
\(B\)-splines of degree \(r\geq \lceil\beta\rceil-1\) on quasi-uniform
knot sequences
\cite[Ch.~6, Corollary~6.21 and Theorem~6.27]{schumaker2007spline}
\cite[Ch.~12, Theorems~12.7--12.8]{schumaker2007spline}.
For piecewise-polynomial finite-element spaces of degree
\(r\geq \lceil\beta\rceil-1\) on shape-regular quasi-uniform meshes of
mesh width \(h\asymp J^{-1}\), the corresponding estimate follows from
the Bramble--Hilbert lemma and the standard finite-element interpolation
estimate
\cite[Theorem~3.1.6]{ciarlet1978finite}
\cite[Ch.~4, Theorem~4.4.20]{brennerScott2008finite}.

For the constructions above, we have 
\(
    \dim(\mathcal S_J)\asymp J^d .
\)
Consequently, after indexing the spaces by their dimension
\(\dsieve=\dim(\mathcal S_J)\), \eqref{eq:examples-Jackson-resolution}
takes the form
\begin{equation}
\label{eq:examples-Jackson-dimension}
    \inf_{v\in\mathcal S_{\dsieve}}
    \|f-v\|_{\infty}
    \leq
    C \dsieve^{-\beta/d}
    \|f\|_{\mathcal C^\beta(\mathsf S)} .
\end{equation}

Let 
\(
    \Pi_{\mathcal V_{\dsieve}}:\mathbb H\to\mathcal V_{\dsieve},
 \) and 
\(    \Pi_{\mathcal W_{\dsieve}}:\mathbb G\to\mathcal W_{\dsieve}
\) denote the corresponding orthogonal projections.
The Galerkin approximation of \(\MKQ\)  is
defined by
\begin{equation}
\label{eq:Galerkin-Pk-Mk}
    \MKQ_{\dsieve}
    :=
    \Pi_{\mathcal W_{\dsieve}}
    \MKQ
    \Pi_{\mathcal V_{\dsieve}},
\end{equation}
\begin{lemma}
\label{lem:Galerkin-approximation-rate}
Under \Cref{assum:kernel_approximation,assum:kernel_smoothness}, 
\begin{equation}
\label{eq:Galerkin-approximation-rate}
    \|\MKQ-\MKQ_{\dsieve}\|_{\mathbb H\to\mathbb G}
    \leq 2 L_{\kappa} C_{\mathcal V}\,
    \dsieve^{-\beta/d}.
\end{equation}
\end{lemma}
\begin{proof}
    The proof is given in Appendix~\ref{app:proof of galerkin rate}.
\end{proof}

Let \(\mathcal M_{\dsieve}
    :=
    \mathcal I-\gamma\MKQ_{\dsieve}.\) Since $\MKQ$, $\Pi_{\mathcal V_\dsieve}$, and $\Pi_{\mathcal W_\dsieve}$ are contractions, $\|\MKQ_\dsieve\|\leq1$. Hence, for every $f\in\mathbb H$,
\begin{equation}
\label{eq:Mk-lower-bound}
    \|\mathcal M_\dsieve f\|_{\mathbb G}
    \geq
    (1-\gamma)\|f\|_{\mathbb H}.
\end{equation}
Thus $\mathcal M_\dsieve$ is injective and $\operatorname{Ran}(\mathcal M_\dsieve)$ is closed. Define
\begin{equation}
\label{eq:intermediate-rewards}
    \widehat R_\dsieve
    :=
    (\operatorname{Id}_{\mathbb G}-\Pi_{\mathcal M_\dsieve})\widehat r.
\end{equation}

We use the following perturbation bound for orthogonal projectors onto operator ranges.
\begin{lemma}
\label{lem:range-projector}
Let \(A,B:\mathbb H\to\mathbb G\) be linear operators satisfying
\(
    \|Af\|_{\mathbb G}\geq\sigma_A\|f\|_{\mathbb H},\) and \(
    \|Bf\|_{\mathbb G}\geq\sigma_B\|f\|_{\mathbb{H}}\,
\) 
 for some \(\sigma_A,\sigma_B>0\) and all \(f\in\mathbb H\). Let \(\Pi_A\) and \(\Pi_B\) be the orthogonal projectors onto their ranges. Then
\begin{equation}
\label{eq:range-projector-perturbation}
    \|\Pi_A-\Pi_B\|
    \leq
    \|A-B\|
    \left(
        \frac{1}{\sigma_A}
        +
        \frac{1}{\sigma_B}
    \right).
\end{equation}
\end{lemma}
\begin{proof}
    The proof is given in Appendix~\ref{app:proof-range-projector}.
\end{proof}
For more general perturbation results for closed subspaces and their associated
orthogonal projectors, see also
\cite[Theorem~I.6.34]{kato1958perturbation} and
\cite[Lemma~221]{kato1966perturbation}.

Applying \Cref{lem:range-projector} to \(\mathcal M\) and \(\mathcal M_\dsieve\), and using \eqref{eq:M-lower-bound}--\eqref{eq:Mk-lower-bound}, gives
\begin{equation}
\label{eq:projector-M-Mk-bound}
    \|\Pi_{\mathcal M}-\Pi_{\mathcal M_\dsieve}\|
    \leq
    \frac{2\gamma}{1-\gamma}
    \|\MKQ-\MKQ_\dsieve\|.
\end{equation}
Consequently, under \Cref{assum: F} and \Cref{lem:Galerkin-approximation-rate},
\begin{align}
\label{eq:Galerkin-reward-term}
    \|\widehat R-\widehat R_\dsieve\|_{\rho^\star}
    & = \|(\Pi_{\mathcal M} - \Pi_{\mathcal{M}_\dsieve})\widehat{r}\|_{\rho^\star}\le
    \frac{2\gamma}{1-\gamma}2L_{\kappa}\mathcal{C}_{\mathcal{V}} \,\dsieve ^{-\beta/d}\lambda 
    \log\frac{1}{\alpha}
    \eqsp.
\end{align}

\subsection{Two-timescale representation of the normal equations}
\label{subsec:finite-dimensional-projection}
We emphasize that \Cref{assum:kernel_approximation} alone does not yield a finite-dimensional representation of the reward estimator \(\widehat R_\dsieve\). To reduce the estimation of \(\widehat R_\dsieve\) to that of a finite-dimensional parameter vector, an additional decomposition is required.

By \eqref{eq:Mk-lower-bound}, there is a unique \(f_\dsieve^\star\in\mathbb H\) such that
\begin{equation}
\label{eq:fk-projected-definition}
    \mathcal M_\dsieve f_\dsieve^\star
    =
    -\Pi_{\mathcal M_\dsieve}\widehat r,
    \qquad
    \widehat R_\dsieve
    =
    \widehat r+\mathcal M_\dsieve f_\dsieve^\star.
\end{equation}
Since \(\mathbb H=\mathcal V_\dsieve\oplus\mathcal V_\dsieve^\perp\),  the function \(f_\dsieve^\star\) admits the unique decomposition
\[
    f^\star_\dsieve=\Phi_\dsieve\theta_\dsieve^\star+h_\dsieve,
    \qquad
    \theta_\dsieve^\star\in\mathbb R^\dsieve,
    \quad
    h_\dsieve\in\mathcal V_\dsieve^\perp.
\]
Introduce the orthogonal projector \(\mathcal Q_\dsieve\) onto \(\mathcal I(\mathcal V_\dsieve^\perp)^\perp\):
\begin{equation}
\label{eq:Qk-projector}
    \mathcal Q_\dsieve
    :=
    \operatorname{Id}_{\mathbb G}
    -
    \mathcal I
    (\operatorname{Id}_{\mathbb H}-\Pi_{\mathcal V_\dsieve})
    \mathcal I^\ast.
\end{equation}
\begin{lemma}
\label{lem:theta_equation}
 The coefficient vector \(\theta_\dsieve^\star\) is the unique solution of the normal equations
\begin{equation}
\label{eq:normal-equations-operator}
    \Phi_\dsieve^\ast\mathcal M_\dsieve^\ast\mathcal Q_\dsieve
    \bigl(
        \widehat r+\mathcal M_\dsieve\Phi_\dsieve\theta_\dsieve^\star
    \bigr)
    =0.
\end{equation}
Moreover, the corresponding reward function admits the representation
\begin{equation}
\label{eq:Rk-Qk-form}
    \widehat R_\dsieve
    =
    \mathcal Q_\dsieve
    \bigl(
        \widehat r+\mathcal M_\dsieve\Phi_\dsieve\theta_\dsieve^\star
    \bigr),
\end{equation}
\end{lemma}
\begin{proof}
    The proof is given in Appendix~\ref{app:theta_equation}.
\end{proof}
\Cref{lem:theta_equation} characterizes \(\theta_\dsieve^\star\) as the unique solution of the linear system \eqref{eq:normal-equations-operator}. However, this characterization is still not sufficient to reduce the problem to a standard single-timescale stochastic approximation scheme. First, the coefficient matrix in \eqref{eq:normal-equations-operator} does not admit unbiased one-sample estimators from the observed trajectory. Second, estimating \(\theta_\dsieve^\star\) alone is not enough to recover the corresponding reward estimator \(\widehat R_\dsieve\), since the representation \eqref{eq:Rk-Qk-form} still involves the unknown operators \(\mathcal Q_\dsieve\) and \(\mathcal M_\dsieve\). These two difficulties naturally lead to a two-timescale stochastic approximation framework \cite{butyrin2026gaussian}, in which auxiliary fast variables are introduced to decouple the estimation tasks and to obtain stochastic updates based on the observed trajectory. The slow iterate tracks the target parameter \(\theta_\dsieve^\star\), while the fast iterates estimate the auxiliary quantities required to represent the normal equations and reconstruct the reward estimator.

Introduce the following cross-moment matrices, which collect the inner products associated with the sieve bases \(\Phi_\dsieve\) and \(\Psi_\dsieve\):
\begin{equation}
\label{eq:population-moments-1}
\begin{aligned}
    &G_{\phi,\dsieve}
    :=
    \Phi_\dsieve^\ast\Phi_\dsieve\in \rset^{\dsieve\times\dsieve},
    \qquad&
    B_\dsieve
    &:=
    \Psi_\dsieve^\ast\MKQ\Phi_\dsieve\in \rset^{\dsievepsi\times\dsieve},
    \\
    &G_{\psi,\dsieve}
    :=
\Psi_\dsieve^\ast\Psi_\dsieve\in \rset^{\dsievepsi\times\dsievepsi},
    &
    C_\dsieve
    &:=
    \Psi_\dsieve^\ast\mathcal I\Phi_\dsieve\in \rset^{\dsievepsi\times\dsieve},\\
    &H_\dsieve
    :=
    \Psi_\dsieve^\ast
    (\operatorname{Id}_{\mathbb G}-\mathcal I\mathcal I^\ast)
    \Psi_\dsieve\in \rset^{\dsievepsi\times\dsievepsi}.
\end{aligned}
\end{equation}
We also define the corresponding vectors
\begin{align}
\label{eq:population-moments-2}
    q_\dsieve
    &:=
    \Phi_\dsieve^\ast\mathcal I^\ast\widehat r\in \rset^\dsieve,
    &
    d_\dsieve
    &:=
    \Psi_\dsieve^\ast
    (\operatorname{Id}_{\mathbb G}-\mathcal I\mathcal I^\ast)
    \widehat r\in\rset^{\dsievepsi}\eqsp,
\end{align}
and exact conditional means
\begin{equation}
\label{eq:exact-centers}
    \overline\Psi_\dsieve^\star(s)
    :=
    \sum_{a\in\mathsf A}
    \pi^\star(a\mid s)\Psi_\dsieve(s,a),
    \qquad
    \overline{r}^{\,\star}(s)
    :=
    \sum_{a\in\mathsf A}
    \pi^\star(a\mid s)\widehat r(s,a).
\end{equation}
The normal equation \eqref{eq:normal-equations-operator} is equivalent to the existence of  \(z^\star_\dsieve\in\rset^{\dsieve}\) and  \(u_\dsieve^\star, v^\star_\dsieve\in\rset ^{\dsieve|\mathsf{A}|}\) satisfying
\begin{equation}
\label{eq:coupled-system}
\begin{cases}
    G_{\psi,\dsieve}u_\dsieve^\star-B_\dsieve\theta_\dsieve^\star
    =0,
    \\[0.3em]
    G_{\phi,\dsieve}z_\dsieve^\star
    -
    \gamma C_\dsieve^\top u_\dsieve^\star
    +
    q_\dsieve
    =0,
    \\[0.3em]

    G_{\psi,\dsieve}v_\dsieve^\star
    +
    \gamma H_\dsieve u_\dsieve^\star
    +
    C_\dsieve z_\dsieve^\star
    -
    C_\dsieve \theta_\dsieve^\star
    -
    d_\dsieve
    =0,
    \\[0.3em]
    G_{\phi,\dsieve}\theta_\dsieve^\star
    -
    \gamma C_\dsieve^\top u_\dsieve^\star
    -
    \gamma B_\dsieve^\top v_\dsieve^\star
    +
    q_\dsieve
    =0.
    \\[0.3em]

\end{cases}
\end{equation}
Moreover, the Galerkin reward estimator \(\widehat R_\dsieve\) admits the representation
\begin{equation}
\label{eq:Rk-coordinate-representation}
    \widehat R_\dsieve(s,a)
    =
    \widehat r(s,a)-\overline{r}^{\,\star}(s)
    -
    \gamma
    \bigl(
        \Psi_\dsieve(s,a)-\overline\Psi_\dsieve^\star(s)
    \bigr)^\top u_\dsieve^\star
    +
    \Phi_\dsieve(s)^\top
    (\theta_\dsieve^\star-z_\dsieve^\star).
\end{equation}
The pair \(
    \theta_{\dsieve}^\star\), 
    \(
      w^\star_{\dsieve} = (u_\dsieve^\star,z_\dsieve^\star,v_\dsieve^\star)
\) is characterized as the unique solution of the linear system
\begin{equation}
\label{eq:TTSA-population-form}
    A_{11,\dsieve}\theta+A_{12,\dsieve}w=b_{1,\dsieve},
    \qquad
    A_{21,\dsieve}\theta+A_{22,\dsieve}^{}w=b_{2,\dsieve}^{}\eqsp,
\end{equation}
where the associated population matrices and vectors are given by

\begin{equation}
\label{eq:TTSA-population-matrix}
A_{\dsieve}^{}
:=
\left(
\begin{array}{c:c}
\underbrace{
    \begin{pmatrix}
    G_{\phi,\dsieve}
    \end{pmatrix}
}_{\scriptstyle A_{11,\dsieve}}
&
\underbrace{
    \begin{pmatrix}
        -\gamma C_\dsieve^\top
        &
        0_{\dsieve\times \dsieve}
        &
        -\gamma B_\dsieve^\top
    \end{pmatrix}
}_{\scriptstyle A_{12,\dsieve}}
\\[1.2em]
\hdashline
\\[-0.4em]
\underbrace{
    \begin{pmatrix}
        -B_\dsieve
        \\[0.3em]
        0_{\dsieve\times \dsieve}
        \\[0.3em]
        -C_\dsieve
    \end{pmatrix}
}_{\scriptstyle A_{21,\dsieve}}
&
\underbrace{
    \begin{pmatrix}
        G_{\psi,\dsieve}
        &
        0_{\dsievepsi\times \dsieve}
        &
        0_{\dsievepsi\times \dsievepsi}
        \\
        -\gamma C_\dsieve^\top
        &
        G_{\phi,\dsieve}
        &
        0_{\dsieve\times \dsievepsi}
        \\
        \gamma H_\dsieve^{}
        &
        C_\dsieve
        &
        G_{\psi,\dsieve}
    \end{pmatrix}
}_{\scriptstyle A_{22,\dsieve}^{}}
\end{array}
\right), \qquad 
b_\dsieve^{}
:=
\left(
\begin{array}{c}
\underbrace{
    -q_\dsieve
}_{\scriptstyle b_{1,\dsieve}}
\\[1.0em]
\hdashline
\\[-0.4em]
\underbrace{
    \begin{pmatrix}
        0_{\dsievepsi}
        \\
            -q_\dsieve
        \\
        d_\dsieve^{}
    \end{pmatrix}
}_{\scriptstyle b_{2,\dsieve}}
\end{array}
\right).
\end{equation}
A detailed derivation of
\eqref{eq:coupled-system} and \eqref{eq:Rk-coordinate-representation}
is provided in Appendix~\ref{app:coupled-system-equivalence}.
We note that each unknown population matrix appearing in
\eqref{eq:coupled-system} admits a  one-sample estimator based on
\((S_t,A_t,S_{t+1})\), while \eqref{eq:Rk-coordinate-representation} provides an exact representation of \(\widehat R_\dsieve\).

\subsection{Plug-in conditional centering}
\label{subsec:plugin-centering}
Note that unbiased estimators of  \(d_\dsieve\) and  \(H_\dsieve\) are still unavailable because the exact conditional means in \eqref{eq:exact-centers} depend on the unknown expert policy \(\pi^\star\). We therefore replace them by plug-in conditional means constructed from the policy estimator \(\widehat\pi\) obtained at the behavior-cloning stage in  \eqref{hat pi def}. 
Specifically, we define the \(\widehat\pi\)-conditional means of the sieve basis and the reward by
\begin{equation}
\label{eq:plugin-centers}
    \overline\Psi_\dsieve^{\widehat\pi}(s)
    :=
    \sum_{a\in\mathsf A}
    \widehat\pi(a\mid s)\Psi_\dsieve(s,a),
    \qquad
    \overline{r}^{\,\widehat\pi}(s)
    :=
    \sum_{a\in\mathsf A}
    \widehat\pi(a\mid s)\widehat r(s,a).
\end{equation}
These quantities provide implementable counterparts of the conditional means in \eqref{eq:exact-centers}. 
\begin{align}
\label{eq:plugin-H-d}
    H_\dsieve^{\widehat\pi}
    &:=
    \mathbb E_{\rho^\star}
    \left[
        \big(\Psi_\dsieve-\overline\Psi_\dsieve^{\widehat\pi}\big)
        \big(\Psi_\dsieve-\overline\Psi_\dsieve^{\widehat\pi}\big)^\top
    \right],\quad
    d_\dsieve^{\widehat\pi}
    :=
    \mathbb E_{\rho^\star}
    \left[
        \big(\Psi_\dsieve-\overline\Psi_\dsieve^{\widehat\pi}\big)
        \widehat r
    \right].
\end{align}
Let \(A_{22,\dsieve}^{\widehat\pi}\) and \(b_{2,\dsieve}^{\widehat\pi}\) denote, respectively, the matrix \(A_{22,\dsieve}\) and the vector \(b_{2,\dsieve}\) with \(H_\dsieve\) and \(d_\dsieve\) replaced by \(H_\dsieve^{\widehat\pi}\) and \(d_\dsieve^{\widehat\pi}\). Let
\(
    \theta_{\dsieve,\widehat\pi}^\star,
    w^\star_{\dsieve,\widehat\pi}
    =
    \bigl(
        u_{\dsieve,\widehat\pi}^\star,
        z_{\dsieve,\widehat\pi}^\star,
        v_{\dsieve,\widehat\pi}^\star
    \bigr)
\)
denote the solution of the perturbed linear system
\begin{equation}
\label{eq:TTSA-population-form-perturb}
    A_{11,\dsieve}\theta+A_{12,\dsieve}w=b_{1,\dsieve},
    \qquad
    A_{21,\dsieve}\theta+A_{22,\dsieve}^{\widehat\pi}w
    =
    b_{2,\dsieve}^{\widehat\pi}\eqsp.
\end{equation}
We define the corresponding plug-in reward proxy by
\begin{align}
\label{eq:Rk-plugin-target}
    \widehat R_{\dsieve,\widehat\pi}(s,a)
    &:=
    \widehat r(s,a)-\overline{r}^{\,\widehat\pi}(s)
    -
    \gamma
    \bigl(
        \Psi_\dsieve (s,a)-\overline\Psi_\dsieve^{\widehat\pi}(s)
    \bigr)^\top u_{\dsieve,\widehat\pi}^\star
    \\
    &\qquad+\Phi_\dsieve (s)^\top
    (\theta_{\dsieve,\widehat\pi}^\star-z_{\dsieve,\widehat\pi}^\star).
\end{align}
For convenience, denote the right-hand side of \eqref{eq:hp-reward} by
\begin{align}
\label{eq:E_frak}
    \mathfrak E_N(\delta)
    &:=
    \sqrt{\Delta(\mathcal F)}
    +
    \sqrt{
        \frac{\mathfrak C_N(\delta)\tau_{\mathrm{mix}}}{N}
    }.
\end{align}
\begin{proposition}
\label{prop:plugin-centering-error}
Let \Cref{assum: MDP,assum: F,assum:UGE,assum:kernel_smoothness,assum:kernel_approximation} hold and $N\geq2\tmix$. For $\delta\in(0,1)$, with probability at least $1-\delta$,
\begin{equation}
\label{eq:plugin-centering-error-bound}
\|
        \widehat R_{\dsieve,\widehat\pi}
        -
        \widehat R_\dsieve
    \|_{\rho^\star}
    \leq
    \frac{\mathfrak C_{\phi,\dsieve}\lambda \log(\frac{1}{\alpha})}{(1-\gamma)^4}
    \left(
        {\mathfrak E_N(\delta)}
        +
        \mathfrak E_N^2(\delta)
    \right).
\end{equation}
where \(\mathfrak C_{\phi, \dsieve}
=
\mathfrak{B}_{\phi,\dsieve}^{16}\lambda_{\min}^{-4}(G_{\phi,\dsieve})\lambda_{\min}^{-4}(G_{\psi,\dsieve})\).
\end{proposition}
\begin{proof}
The proof is given in Appendix~\ref{app:proof-plugin-centering}.
\end{proof}

\Cref{prop:plugin-centering-error} quantifies the perturbation induced by replacing the exact conditional means by their plug-in counterparts based on \(\widehat\pi\). In particular, it controls the resulting perturbation of the solution to \eqref{eq:coupled-system} and, consequently, the deviation of the plug-in reward proxy \(\widehat R_{\dsieve,\widehat\pi}\) in \eqref{eq:Rk-plugin-target} from the exact Galerkin reward estimator \(\widehat R_\dsieve\) in \eqref{eq:Rk-coordinate-representation}.

\subsection{Two-timescale stochastic approximation}
\label{subsec:TTSA-algorithm}

Recall that the behavior-cloning stage constructs the policy estimator $\widehat\pi$ from the expert trajectory
\(
    \mathcal D:=\{(S_i,A_i)\}_{i=1}^{N},
\)
To approximate \(\widehat R_{\dsieve, \widehat \pi}\) in \eqref{eq:Rk-plugin-target}, we introduce a second, independent expert trajectory
\(
    \mathcal D'
    :=
    \{(S_n',A_n')\}_{n=1}^{N},
\) of the same size
generated under the expert policy $\pi^\star$. The independence of $\mathcal D$ and $\mathcal D'$ allows us to condition on $\mathcal D$ throughout the second stage and assume that the estimators $\widehat\pi$ and $\widehat r$ are fixed.

Systems of the form \eqref{eq:TTSA-population-form} can be solved efficiently within the two-timescale stochastic approximation (TTSA) framework; see \cite{butyrin2026gaussian}. For completeness, Appendix~\ref{app:ttsa_framework} collects the formal definition of the TTSA recursion, the assumptions required for its convergence, and the corresponding finite-time convergence guarantees.
Condition on \(\mathcal D\). For \(n=1,\ldots,N\), set \(\widehat r_n:=\widehat r(S_n',A_n')\) and 
\begin{align}
\label{eq:sample-features}
\phi_n:=\Phi_\dsieve(S_n'),
\quad
\psi_n:=\Psi_\dsieve(S_n',A_n'),
\quad
\widetilde\psi_n
    :=
    \psi_n-
\overline\Psi_\dsieve^{\widehat\pi}(S_n').
\end{align}
Let
\(
    w_n
    :=
    (
        u_n,z_n,v_n
    )^{\top}\eqsp.
\) 

Single-transition representatives of \eqref{eq:TTSA-population-matrix} are

\begin{equation}
\label{eq:sample-A}
\funcAw_{\dsieve}^{\widehat\pi}(X_n)
=
\left(
\begin{array}{c:c}
\underbrace{
    \begin{pmatrix}
    \phi_n\phi_n^\top
    \end{pmatrix}
}_{\scriptstyle \funcAw_{11,\dsieve}(X_n)}
&
\underbrace{
    \begin{pmatrix}
        -\gamma\phi_n\psi_n^\top
        &
        0_{\dsieve\times \dsieve}
        &
        -\gamma\phi_{n+1}\psi_n^\top
    \end{pmatrix}
}_{\scriptstyle \funcAw_{12,\dsieve}(X_n)}
\\[1.2em]
\hdashline
\\[-0.4em]
\underbrace{
    \begin{pmatrix}
        -\psi_n\phi_{n+1}^\top
        \\[0.3em]
        0_{\dsieve\times \dsieve}
        \\[0.3em]
        -\psi_n\phi_n^\top
    \end{pmatrix}
}_{\scriptstyle \funcAw_{21,\dsieve}(X_n)}
&
\underbrace{
    \begin{pmatrix}
        \psi_n\psi_n^\top
        &
        0_{\dsievepsi\times \dsieve}
        &
        0_{\dsievepsi\times \dsievepsi}
        \\
        -\gamma\phi_n\psi_n^\top
        &
        \phi_n\phi_n^\top
        &
        0_{\dsieve\times \dsievepsi}
        \\
\gamma\widetilde\psi_n\widetilde\psi_n^\top
        &
        \psi_n\phi_n^\top
        &
        \psi_n\psi_n^\top
    \end{pmatrix}
}_{\scriptstyle \funcAw_{22,\dsieve}^{\widehat\pi}(X_n)}
\end{array}
\right)\eqsp, \,\,\, \funcbw_\dsieve^{\widehat\pi}(X_n)
=
\left(
\begin{array}{c}
\underbrace{
    -\phi_n\widehat r_n
}_{\scriptstyle \funcbw_{1,\dsieve}(X_n)}
\\[1.0em]
\hdashline
\\[-0.4em]
\underbrace{
    \begin{pmatrix}
        0_{\dsievepsi}
        \\
        -\phi_n\widehat r_n
        \\
        \widetilde\psi_n\widehat r_n
    \end{pmatrix}
}_{\scriptstyle \funcbw_{2,\dsieve}^{\widehat\pi}(X_n)}
\end{array}
\right)\eqsp,
\end{equation}
where \(X_n = (S_n', A_n', S_{n+1}').\)
Let $(\beta_k)_{k \geq 0}$ and $(\gamma_k)_{k \geq 0}$ be non-increasing sequences of the form 
\begin{equation}
\label{eq:poly_steps}
\beta_k = c_{0, \beta} (k+k_0)^{-b}\eqsp,\quad\gamma_k = c_{0, \gamma} (k+k_0)^{-a} \eqsp, \quad \frac{1}{2} < a < b < 1\eqsp.
\end{equation}
The  TTSA recursion is
\begin{align}
\label{eq:TTSA-theta-update}
    \theta_{n+1}
    &=
    \theta_n
    +
    \beta_n
    \left\{
        \funcbw_{1,\dsieve}(X_n)
        -
        \funcAw_{11,\dsieve}(X_n)\theta_n
        -
        \funcAw_{12,\dsieve}(X_n)w_n
    \right\},
    \\
\label{eq:TTSA-w-update}
    w_{n+1}
    &=
    w_n
    +
    \gamma_n
    \left\{
        \funcbw_{2,\dsieve}^{\widehat\pi}(X_n)
        -
        \funcAw_{21,\dsieve}(X_n)\theta_n
        -
        \funcAw_{22,\dsieve}^{\widehat\pi}(X_n)w_n
    \right\}\eqsp,
\end{align}
The reward returned after \(N\) updates is
\begin{align}
\label{eq:TTSA-reward-output}
    \widehat R_{\dsieve}^{\mathrm{SA}}(s,a)
    &:=
    \widehat r(s,a)-\overline{r}^{\,\widehat\pi}(s)
    -
    \gamma
    \bigl(
        \Psi_\dsieve (s,a)-\overline\Psi_\dsieve^{\widehat\pi}(s)
    \bigr)^\top u_{N}
    +
    \Phi_\dsieve(s)^\top(\theta_{N}-z_{N}).
\end{align}
The complete procedure for constructing
\(\widehat R_{\dsieve}^{\mathrm{SA}}\) is summarized in
\Cref{alg:ttsa}.
\begin{algorithm}[H]
\caption{Two-timescale estimation of the Galerkin Bellman projection}
\label{alg:ttsa}
\begin{algorithmic}[1]
\Require \(\mathcal D'\), \(\widehat\pi\), \(\widehat r=\lambda\log\widehat\pi\), bases \(\Phi_\dsieve,\Psi_\dsieve\) and step sizes \(\{\beta_n,\gamma_n\}_{n=0}^{N-1}\).
\State Set initial vectors \(\theta_0=0\) and \(w_0 =0 \).
\For{\(n=0,\ldots,N-1\)}
    \State Form \(\phi_n,\phi_{n+1},\psi_n,\widehat r_n,\widetilde\psi_n\) using \eqref{eq:sample-features}.
    \State Form \(\funcAw_{ij,\dsieve}(X_n)\) and \(\funcbw_{i,\dsieve}(X_n)\) using \eqref{eq:sample-A}.
    \State Update \(\theta_{n+1}\) and \(w_{n+1}\) simultaneously using \eqref{eq:TTSA-theta-update}--\eqref{eq:TTSA-w-update}.
\EndFor
\State \Return \(\widehat R_{\dsieve}^{\mathrm{SA}}\) defined by \eqref{eq:TTSA-reward-output}.
\end{algorithmic}
\end{algorithm}
We next establish a finite-sample bound for the estimator
\(\widehat R_{\dsieve}^{\mathrm{SA}}\) in \eqref{eq:TTSA-reward-output}.
Throughout this result, the step sizes are chosen according to
\eqref{eq:poly_steps} with \(a=1-\frac{1}{\log N}\) and a sufficiently small
initial stepsize ratio.

\begin{proposition}
\label{prop:TTSA-error-bound}
Suppose that \Cref{assum: MDP,assum: F,assum:UGE,assum:kernel_smoothness,assum:kernel_approximation} hold. Then, conditionally on \(\mathcal D\), with probability at least \(1-\delta\),
\begin{equation}
\label{eq:reward-TTSA-high-probability-rate}
    \|\widehat R_{\dsieve}^{\mathrm{SA}}
      -\widehat R_{\dsieve,\widehat\pi}\|_{\rho^\star}
    \le
    \frac{C_{\mathrm{SA},\dsieve}\log^3(\frac{2}{\delta})}{\sqrt{N}}.
\end{equation}
\end{proposition}
\begin{proof}
The proof is given in Appendix~\ref{app:ttsa_rate}.
\end{proof}

\subsection{Final result}
\label{subsec:estimated-Bellman-final}

\begin{theorem}
\label{thm:estimated-Bellman-operator}
Suppose that
\Cref{assum: MDP,assum: F,assum:UGE,assum:kernel_smoothness,assum:kernel_approximation}
hold. Let \(N\ge 2\tmix\) and \(\widehat R_{\dsieve}^{\mathrm{SA}}\) be the output of
\Cref{alg:ttsa} with the stepsizes specified in
\Cref{prop:TTSA-error-bound}. Then, with probability at least \(1-\delta\),
\begin{align}
\label{eq:final-reward-bound}
    \|R_{\mathrm{LS}}^\star-\widehat R_{\dsieve}^{\mathrm{SA}}\|_{\rho^\star}
    \lesssim\;&
    \lambda
    \sqrt{\log(1/\alpha) + 2}\,
    \mathfrak E_N(\delta/2)
    +
    \frac{
        L_{\kappa}C_{\mathcal V}
        \lambda\log(1/\alpha)
    }{1-\gamma}
    \dsieve^{-\beta/d}
    \\
    &+
    \frac{
        \mathfrak C_{\phi,\dsieve}
        \lambda\log(1/\alpha)
    }{(1-\gamma)^4}
    \left(
        \mathfrak E_N(\delta/2)
        +
        \mathfrak E_N^2(\delta/2)
    \right)
    +
    \frac{
        C_{\mathrm{SA},\dsieve}
        \log^3(4/\delta)
    }{\sqrt N}\eqsp,
\end{align}
where \(\mathfrak E_N(\delta/2)\) is defined in \eqref{eq:E_frak}.
\end{theorem}

\begin{proof}
We use the decomposition in \eqref{eq:four-term-decomposition} and bound each term separately.
By \Cref{thm:hp-reward}, with probability at least \(1-\delta/2\),
\begin{equation}
\label{eq:KL-rate-section5}
    \sqrt{\mathcal K(\widehat\pi)}
    \lesssim
    \mathfrak E_N(\delta/2),
    \qquad
    \|R_{\mathrm{LS}}^\star-\widehat R\|_{\rho^\star}
    \lesssim
    \lambda
    \left(\log\frac1\alpha+2\right)^{1/2}
    \mathfrak E_N(\delta/2).
\end{equation}
The Galerkin approximation bound \eqref{eq:Galerkin-reward-term} gives
\begin{equation}
\label{eq:final-Galerkin-term}
    \|\widehat R-\widehat R_\dsieve\|_{\rho^\star}
    \leq
    \frac{
        4L_{\kappa} C_{\mathcal V}
        \lambda\log(1/\alpha)
    }{1-\gamma}
    \dsieve^{-\beta/d}.
\end{equation}
Moreover, on the event in \eqref{eq:KL-rate-section5},
\Cref{prop:plugin-centering-error} yields
\begin{equation}
\label{eq:final-plugin-term}
    \|\widehat R_\dsieve-\widehat R_{\dsieve,\widehat\pi}\|_{\rho^\star}
    \lesssim
    \frac{
        \mathfrak C_{\phi,\dsieve}
        \lambda\log(1/\alpha)
    }{(1-\gamma)^4}
    \left(
        \mathfrak E_N(\delta/2)
        +
        \mathfrak E_N^2(\delta/2)
    \right).
\end{equation}
Finally, conditionally on \(\mathcal D\), \Cref{prop:TTSA-error-bound} implies
that, with probability at least \(1-\delta/2\),
\begin{equation}
\label{eq:final-TTSA-term}
    \|\widehat R_{\dsieve}^{\mathrm{SA}}
      -\widehat R_{\dsieve,\widehat\pi}\|_{\rho^\star}
    \lesssim
    \frac{
        C_{\mathrm{SA},\dsieve}
        \log^3(4/\delta)
    }{\sqrt N}.
\end{equation}
Combining
\eqref{eq:KL-rate-section5}--\eqref{eq:final-TTSA-term}
and applying the union bound proves \eqref{eq:final-reward-bound}.
\end{proof}

\bibliographystyle{emss}
\bibliography{refs}

\appendix

\section{Proof of Theorem \ref{thm:hp-reward}}
\label{app:proof-oracle}

\subsection{Proof of Proposition \ref{prop:reward shaping lemma}}
\label{app:compatible}
\begin{proof}
Fix 
\(
f\in L_2(d_{\pi^\star})\) 
and set
\[
    V_f(s):=f(s)\eqsp, \quad 
Q_f(s,a):=f(s)+\lambda\log\pi^\star(a\mid s).
\]
By \eqref{eq:R_f},
\(
    R_f+\gamma\MKQ V_f
    =
    Q_f.
\)
Moreover,
\[
\begin{aligned}
    \lambda\log\sum_{a\in\mathsf A}
    \exp\!\left({Q_f(s,a)}/{\lambda}\right)
    &=
    \lambda\log\Big(
        \exp\big({f(s)/\lambda}\big)
        \sum_{a\in\mathsf A}\pi^\star(a\mid s)
    \Big)
    =f(s)=V_f(s).
\end{aligned}
\]
Thus \(Q_f\) and \(V_f\) satisfy the soft Bellman optimality equations,
and their associated softmax policy is
\[
    \frac{\exp(Q_f(s,a)/\lambda)}
    {\sum_{a'}\exp(Q_f(s,a')/\lambda)}
    =
    \pi^\star(a\mid s).
\]

Conversely, let \(R\) be compatible with \(\pi^\star\). Since
\(\pi_R^\star=\pi^\star\), the softmax identity and
\eqref{eq:soft-bellman} imply
\[
    Q_R^\star(s,a)-V_R^\star(s)
    =
    \lambda\log\pi^\star(a\mid s).
\]
Combining this with
\(Q_R^\star=R+\gamma\MKQ V_R^\star\) yields
\[
    R(s,a)
    =
    V_R^\star(s)
    +
    \lambda\log\pi^\star(a\mid s)
    -
    \gamma\MKQ V_R^\star(s,a),
\]
which is \eqref{eq:R_f} with \(f=V_R^\star\).
Finally, \(R\in L_2(\rho^\star)\), stationarity of
\(d_{\pi^\star}\), and the fact that the entropy term is bounded imply
\(V_R^\star\in L_2(d_{\pi^\star})\).
\end{proof}

\subsection{Proof of Theorem \ref{thm:hp-reward}}
\label{subsec:policy-oracle-proof}
If \(\Delta(\mathcal F)=+\infty\), the assertion is trivial. Hence,
assume \(\Delta(\mathcal F)<\infty\). By \eqref{eq: L def} and the definition of  \(
    \mathcal K(\pi)\)
in \eqref{eq:kl_pi}, 

\[
    \,L(\pi)-L(\pi^\star)
    =
    \mathcal K(\pi).
\]
By compactness of \(\mathcal F\) (see \Cref{rem:compact}), the infimum in
\eqref{eq:Delta_F} is attained. Fix
\[
    \bar\pi
    \in
    \argmin_{\pi\in\mathcal F}
    \left\{
        \mathcal K(\pi)+\mathcal R(\pi)
    \right\}.
\]
Thus,
\begin{equation}
\label{eq:oracle-comparator}
    \mathcal K(\bar\pi)+\mathcal R(\bar\pi)
    =
    \Delta(\mathcal F),
    \qquad
    \mathcal K(\bar\pi)
    \leq
    \Delta(\mathcal F).
\end{equation}
For \(\pi\in\mathcal F\), define
\[
    g_\pi(s,a)
    :=
    \log\frac{\bar\pi(a\mid s)}{\pi(a\mid s)}.
\]
The optimality of \(\widehat\pi\) in \eqref{hat pi def} implies
\begin{equation}
\label{eq:oracle-ERM-inequality}
    \frac1N
    \sum_{n=1}^N
    \log
    \frac{
        \bar\pi(A_n\mid S_n)
    }{
        \widehat\pi(A_n\mid S_n)
    }
    \leq
    \mathcal R(\bar\pi)-\mathcal R(\widehat\pi).
\end{equation}
On the other hand,
\[
    \log\frac{\pi^\star}{\widehat\pi}
    =
    \log\frac{\pi^\star}{\bar\pi}
    +
    \log\frac{\bar\pi}{\widehat\pi},
\]
and therefore
\[
   \mathcal K(\widehat\pi)
    =
    \mathcal K(\bar\pi)
    +
    \E_{\rho^\star}
    \log
    \frac{
        \bar\pi(A\mid S)
    }{
        \widehat\pi(A\mid S)
    }.
\]
Combining this identity with
\eqref{eq:oracle-ERM-inequality} and
\Cref{prop:uniform-likelihood-deviation}, evaluated at
\(\pi=\widehat\pi\), gives
\begin{align}
\label{eq:oracle-self-bound-general}
    \mathcal K(\widehat\pi)
    &\leq
    \mathcal K(\bar\pi)
    +
    \mathcal R(\bar\pi)
    -
    \mathcal R(\widehat\pi)
    \nonumber\\
    &\quad+
    8
    \sqrt{
        \frac{
            2C_\alpha\tau_{\mathrm{mix}}
            \left[
                \mathcal K(\widehat\pi)
                +
                \mathcal K(\bar\pi)
                +
                \varepsilon
            \right]
            M(\varepsilon,\delta)
        }{N}
    }
    \nonumber\\
    &\quad+
    \frac{
        160\log(1/\alpha)\tau_{\mathrm{mix}}
        M(\varepsilon,\delta)
    }{N}
    +
    2\varepsilon.
\end{align}
Since \(\mathcal R(\widehat\pi)\geq0\) and
\eqref{eq:oracle-comparator} holds,
\begin{align}
\label{eq:oracle-self-bound}
    \mathcal K(\widehat\pi)
    &\leq
    \Delta(\mathcal F)
    +
    8
    \sqrt{
        \frac{
            2C_\alpha\tau_{\mathrm{mix}}
            \left[
                \mathcal K(\widehat\pi)
                +
                \Delta(\mathcal F)
                +
                \varepsilon
            \right]
            M(\varepsilon,\delta)
        }{N}
    }
    \nonumber\\
    &\quad+
    \frac{
        160\log(1/\alpha)\tau_{\mathrm{mix}}
        M(\varepsilon,\delta)
    }{N}
    +
    2\varepsilon.
\end{align}
Choose \(\varepsilon_0
    :=
    {
        \Lambda\tau_{\mathrm{mix}}
    }/(
        \Lambda N+\tau_{\mathrm{mix}}
    )\in (0,\Lambda]
\).
Note that
\(
    \varepsilon_0
    \leq
    {\tau_{\mathrm{mix}}}/{N}.
\)
Moreover,
\begin{align*}
    M(\varepsilon_0,\delta)
    =
    D\log\left(
        \frac{
            1+\Lambda N/\tau_{\mathrm{mix}}
        }{\alpha}
    \right)
    +
    \log\left(\frac{\sqrt{2\mathfrak N_0}}{\delta}\right)\eqsp.
\end{align*}
With \(\mathfrak C_N(\delta)\) defined in
\eqref{eq:first-stage-rate}, set
\[
    r_N(\delta)
    :=
    \frac{
        \mathfrak C_N(\delta)\tau_{\mathrm{mix}}
    }{N}
    =
    \frac{
        C_\alpha M(\varepsilon_0,\delta)
        \tau_{\mathrm{mix}}
    }{N}.
\] 
Since \(\mathfrak N_0\geq1\), \(\delta\in(0,1)\), and \(C_\alpha\geq2\),
we have
\(M(\varepsilon_0,\delta)\geq\log\sqrt{2}\). Therefore 
\(
    \varepsilon_0
    \lesssim
    r_N(\delta).
\)
Furthermore,
\[
    \frac{
        \log(1/\alpha)\tau_{\mathrm{mix}}
       M(\varepsilon_0,\delta)
    }{N}
    \leq
    r_N(\delta).
\]
Substituting \(\varepsilon=\varepsilon_0\) into
\eqref{eq:oracle-self-bound}, and using \(\sqrt{x+y+z}
    \leq
    \sqrt x+\sqrt y+\sqrt z,\) we obtain
\begin{equation}
\label{eq:oracle-self-bound-final}
    \mathcal K(\widehat\pi)
    \leq
    \Delta(\mathcal F)
    +
     C\sqrt{r_N(\delta)\mathcal K(\widehat\pi)}
    +
    C\sqrt{r_N(\delta)\Delta(\mathcal F)}
    +
    C r_N(\delta).
\end{equation}
where \(C>0\) is a numerical constant. Young's inequality gives
\(
    C\sqrt{r_N(\delta)\mathcal K(\widehat\pi)}
    \leq
    \frac12\mathcal K(\widehat\pi)
    +
    \frac{C^2}{2}r_N(\delta).
\)
After moving \(\mathcal K(\widehat\pi)\) to the left-hand side, we conclude that
\[
    \mathcal K(\widehat\pi)
    \lesssim
    \Delta(\mathcal F)
    +
    r_N(\delta)
    +
    \sqrt{
        \Delta(\mathcal F)r_N(\delta)
    }.
\]
Equivalently,
\begin{equation}
\label{eq:policy-oracle-final}
    \mathcal K(\widehat\pi)
    \lesssim
    \left(
        \sqrt{\Delta(\mathcal F)}
        +
        \sqrt{
            \frac{
                \mathfrak C_N(\delta)\tau_{\mathrm{mix}}
            }{N}
        }
    \right)^2.
\end{equation}
This proves \eqref{eq:KL-eps}.

\begin{proposition}
\label{prop:likelihood-ratio-variance}
Let \Cref{assum: F} hold. For every \(\pi\in\mathcal F\), 
\[
    \Var_{\rho^\star}\bigl(g_\pi(S,A)\bigr)
    \leq
    2C_\alpha
    \left[
        \mathcal K(\pi)+\mathcal K(\bar\pi)
    \right], 
\]
where \(C_\alpha:=\log\frac1\alpha+2.\)
\end{proposition}

\begin{proof}
Write
\[
    g_\pi(S,A)
    =
    \log\frac{\pi^\star(A\mid S)}{\pi(A\mid S)}
    -
    \log\frac{\pi^\star(A\mid S)}{\bar\pi(A\mid S)}.
\]
Therefore,
\begin{align*}
    \Var_{\rho^\star}\bigl(g_\pi(S,A)\bigr)
\leq
    2\E_{\rho^\star}
    \left[
        \left(
            \log\frac{\pi^\star(A\mid S)}{\pi(A\mid S)}
        \right)^2
    \right]
    +
    2\E_{\rho^\star}
    \left[
        \left(
            \log\frac{\pi^\star(A\mid S)}{\bar\pi(A\mid S)}
        \right)^2
    \right].
\end{align*}
Since every policy in \(\mathcal F\) is bounded below by \(\alpha\),
Proposition~\ref{lem:kl-2}, applied conditionally on \(S\) yields
\[
    \E_{\rho^\star}
    \left[
        \left(
            \log\frac{\pi^\star(A\mid S)}{\pi(A\mid S)}
        \right)^2
    \right]
    \leq
    C_\alpha\mathcal K(\pi),
\]
and the same bound with \(\pi\) replaced by \(\bar\pi\). Combining the
last three displays proves the claim.
\end{proof}

\begin{proposition}
\label{prop:uniform-likelihood-deviation}
Let \Cref{assum: F,assum:UGE} hold and \(N\geq2\tau_{\mathrm{mix}}\). For every
\(\delta\in(0,1)\) and every \(\varepsilon\in(0,\Lambda]\), with
probability at least \(1-\delta\), simultaneously for all
\(\pi\in\mathcal F\),
\begin{align}
\label{eq:uniform-likelihood-deviation}
    \Big|\,
        \E_{\rho^\star}[g_\pi(S,A)]
        -
        \frac1N\sum_{n=1}^N g_\pi(S_n,A_n)
    \,\Big|
    &\leq
     8
    \sqrt{
        \frac{
            2C_\alpha\tau_{\mathrm{mix}}
            \left[
                \mathcal K(\pi)
                +
                \mathcal K(\bar\pi)
                +
                \varepsilon
            \right]
            M(\varepsilon,\delta)
        }{N}
    }\nonumber\\
    &+\frac{
        160\log(1/\alpha)\tau_{\mathrm{mix}}
        M(\varepsilon,\delta)
    }{N}
    +
    2\varepsilon  
\end{align}
where
\(
    M(\varepsilon,\delta)
    :=
    D\log\left(
        \frac{\Lambda}{\alpha\varepsilon}
    \right)
    +
    \log\left(\frac{\sqrt{2\mathfrak N_0}}{\delta}\right).
\)
\end{proposition}

\begin{proof}
Since \(x\mapsto-\log x\) is \(1/\alpha\)-Lipschitz on
\([\alpha,1]\),
\[
    \|g_\pi-g_{\pi'}\|_\infty
    \leq
    \alpha^{-1}\|\pi-\pi'\|_\infty.
\]
Hence an \(\alpha\varepsilon\)-net of \(\mathcal F\) induces an
\(\varepsilon\)-net of
\(
    \mathcal G_{\bar\pi}
    :=
    \{g_\pi:\pi\in\mathcal F\}.
\)
By \eqref{eq:entropy-bound}, there exist
\(\{\pi_j\}_{j=1}^K\subset\mathcal F\) such that
\(
    K
    \leq
    \left(
        {\Lambda}/({\alpha\varepsilon})
    \right)^D
\)
and for every \(\pi\in\mathcal F\) there is \(j\) satisfying
\[
    \|g_\pi-g_{\pi_j}\|_\infty
    \leq
    \varepsilon.
\]
The uniform floor also implies
\(
    \|g_{\pi_j}\|_\infty
    \leq
    \log\frac1\alpha,
\)
and
\(
    \left\|
        g_{\pi_j}
        -
        \E_{\rho^\star}[g_{\pi_j}]
    \right\|_\infty
    \leq
    2\log\frac1\alpha.
\)
Applying Bernstein's inequality,  \Cref{th:bernstein_inequality}  to each \(g_{\pi_j}\),
using \(N\geq2\tau_{\mathrm{mix}}\), and taking a union bound, we obtain,
with probability at least \(1-\delta\), simultaneously for
\(j=1,\ldots,K\),
\begin{align}
\label{eq:net-Bernstein-bound}
    |
        (\E_{\rho^\star} - P_N)g_{\pi_j}  
    |
    &\leq
    \sqrt{
        \frac{
            64\tau_{\mathrm{mix}}
            \Var_{\rho^\star}(g_{\pi_j})
    \log\left(\frac{K\sqrt{2\mathfrak N_0}}{\delta}\right)
        }{N}
    }
    \nonumber\\
    &\quad+
    \frac{
160\log(1/\alpha)\tau_{\mathrm{mix}} 
    \log\left(\frac{K\sqrt{2\mathfrak N_0}}{\delta}\right)
    }{N},
\end{align}
where $P_N g := N^{-1} \sum_{t=1}^N g(S_t, A_t)$. 
For any \(\pi\in\mathcal F\), choose \(j\) such that
\(\|g_\pi-g_{\pi_j}\|_\infty\leq\varepsilon\). Then
\[
    |\mathcal K(\pi_j)-\mathcal K(\pi)|
    =
    \left|
        \E_{\rho^\star}
        \log\frac{\pi(A\mid S)}{\pi_j(A\mid S)}
    \right|
    \leq
    \varepsilon.
\]
Consequently, by
\Cref{prop:likelihood-ratio-variance},
\[
    \Var_{\rho^\star}(g_{\pi_j})
    \leq
    2C_\alpha
    \left[
        \mathcal K(\pi)
        +
        \mathcal K(\bar\pi)
        +
        \varepsilon
    \right].
\]
Moreover,
\[
\begin{aligned}
    |(\E_{\rho^\star}-P_N)g_\pi|
    &\leq
    |(\E_{\rho^\star}-P_N)g_{\pi_j}|
    +
    |\E_{\rho^\star}(g_\pi-g_{\pi_j})|
    +
    |P_N(g_\pi-g_{\pi_j})|
    \\
    &\leq
    |(\E_{\rho^\star}-P_N)g_{\pi_j}|
    +
    2\varepsilon.
\end{aligned}
\]
Substituting these bounds into
\eqref{eq:net-Bernstein-bound} and using
\(
    \log K
    \leq
    D\log\left(
        {\Lambda}/{(\alpha\varepsilon)}
    \right)
\)
proves \eqref{eq:uniform-likelihood-deviation}.
\end{proof}

\begin{proposition}
\label{lem:kl-2}
Fix a state \(s\) and let \(\pi_1(\cdot\mid s),\pi_2(\cdot\mid s)\) be two distributions on a finite action set.
Assume the one-sided bound \(L_s(a):=\tfrac{\pi_1(a\mid s)}{\pi_2(a\mid s)}\le 1/\alpha\) for all \(a\) with \(\pi_1(a\mid s)>0\) (equivalently, \(\pi_2(\cdot\mid s)\ge \alpha\,\pi_1(\cdot\mid s)\)).
Then
\[
\mathbb E_{A \sim \pi_1(\cdot\mid s)}\!\big[(\log L_s(A))^2\big]
\;\le\; \big(\log(1/\alpha)+2\big)\,
\mathrm{KL}\!\big(\pi_1(\cdot\mid s)\mid \pi_2(\cdot\mid s)\big).
\]
\end{proposition}

\begin{proof}
Let \(t>0\) and use the pointwise inequality
\[
\frac{t(\log t)^2}{t\log t - t + 1} \;\le\;
\begin{cases}
2, & t\in(0,1],\\
\log t + 2, & t\ge 1,
\end{cases}
\]
which implies for all \(t\in[0,1/\alpha]\) that
\(\frac{t(\log t)^2}{t\log t - t + 1} \le \log(1/\alpha)+2\).
Set \(t=L_s(a)=\pi_1/\pi_2\). Then, by change of measure,
\[
\mathbb E_{\pi_1}\!\big[(\log L_s)^2\big]
=\mathbb E_{\pi_2}\!\big[L_s(\log L_s)^2\big]
\le \big(\log(1/\alpha)+2\big)\,\mathbb E_{\pi_2}\!\big[L_s\log L_s - L_s + 1\big].
\]
Finally, \(\mathbb E_{\pi_2}[L_s\log L_s - L_s + 1]=\mathrm{KL}(\pi_1\mid \pi_2)\), yielding the claim.
\end{proof}

\paragraph{Bernstein inequality for Markov chains.} We finish this section with Bernstein type inequality for uniformly geometrically ergodic Markov chains. 
Let $\{Z_k\}_{k \in \nset}$ be a Markov chain on $(\Zset,\Zsigma)$ with a Markov kernel $\mathsf K$. We assume that $\mathsf K$ is uniformly geometrically ergodic, that is there exists $\tmix \in \nset$ such that for all $n \in \nset$ and $z \in \Zset$,
\begin{equation}
\label{eq:tv_ergodicity}
\tvnorm{\mathsf K^{n}(z, \cdot) - \pi} \leq (1/4)^{\left\lfloor n/\tmix \right\rfloor}\eqsp.
\end{equation}
There are a number of papers in which Bernstein type inequalities for Markov chains satisfying \eqref{eq:tv_ergodicity} are obtained. We note  \cite{rio2017asymptotic}, \cite{durmus2024probability}, \cite{durmus2023rosenthal}, \cite{paulin2015concentration}. In some of them, the results are formulated in terms of asymptotic variance, and in some in terms of proxy variance, for example, marginal variance $\Var_\pi(f)$. For the purposes of this paper, it is sufficient to have the result in terms of marginal variance.  We will use the result from \cite{paulin2015concentration} in which Bernstein's inequality is obtained with an explicit dependence on all parameters of the Markov chain and with explicit expressions for constants.
\begin{theorem}
\label{th:bernstein_inequality}
Suppose \Cref{assum:UGE} holds and let $f: \Zset \to \rset$ be a bounded measurable function. Let $S_n = \sum_{\ell=1}^n \{f(Z_\ell) - \pi(f)\}$. Then for any $t \geq 0$, 
\begin{equation}
\label{eq:bernstein_inequality}
\PP_{\rho_0}(|S_n| \geq t) \leq \sqrt{2\mathfrak N_0} \exp\biggl\{-\frac{t^2}{32\tmix(n + 2\tmix)\Var_{\pi}(f) + 80 t \tmix \|\barf\|_{\infty}}\biggr\}\eqsp, 
\end{equation}
where $\bar f = f - \pi(f)$. 
\end{theorem}
\begin{proof}
    The proof follows from Theorem~3.11,  Proposition 3.4 and Proposition 3.15 in \cite{paulin2015concentration}. 
\end{proof}

\subsection{Transfer from policy error to reward error}
\label{subsec:proof-reward-stability}

The key point is that the population target and its plug-in analogue use the same orthogonal projector in the same Hilbert space $\mathbb G=L_2(\rho^\star)$:
\[
    R_{\mathrm{LS}}^\star
    =
    (\operatorname{Id}_{\mathbb G}-\Pi_{\mathcal M})r^\star,
    \qquad
    \widehat R
    =
    (\operatorname{Id}_{\mathbb G}-\Pi_{\mathcal M})\widehat r.
\]
Consequently,
\begin{equation}
\label{eq:projection-contraction-reward}
    \|\widehat R-R_{\mathrm{LS}}^\star\|_{\rho^\star}
    = \|(\operatorname{Id}_{\mathbb G}-\Pi_{\mathcal M})(\widehat r - r^\star) \|_{\rho^\star}\leq
    \|\widehat r-r^\star\|_{\rho^\star}.
\end{equation}
Since $\widehat\pi(a\mid s)\geq\alpha$ and $\pi^\star(a\mid s)\leq1$,
\[
    \frac{\pi^\star(a\mid s)}{\widehat\pi(a\mid s)}\leq\frac1\alpha.
\]
Applying Proposition~\ref{lem:kl-2} statewise gives
\begin{align}
    \|\widehat r-r^\star\|_{\rho^\star}^2
    &=
    \lambda^2\E_{\rho^\star}
    \left[
        \left(
            \log\frac{\widehat\pi(A\mid S)}{\pi^\star(A\mid S)}
        \right)^2
    \right]\leq
\lambda^2\left(\log\frac1\alpha+2\right)
    \mathcal{K}(\widehat \pi).
\end{align}
Combining this inequality with \eqref{eq:KL-eps} and using
$2\sqrt{xy}\leq x+y$ yields \eqref{eq:hp-reward} after changing a universal constant.

\section{Proof of Theorem \ref{thm:holder-relu-rate}}
\label{app:proof-holder-relu}
Recall that the optimal softmax policy \(\pi^\star\) is given by \eqref{eq:softmax}.  The approximation quality is measured by the expected conditional KL divergence \(\mathcal K(\pi)\) defined in \eqref{eq:kl_pi}.  We approximate \(\pi^\star\) by first approximating the vector of action logits \(Q^\star(\cdot,\cdot)\) with a neural network.

Let $\sigma(x) = x \vee 0$ be a rectified linear unit (ReLU) activation function. For $v = (v_1, \ldots, v_r) \in \rset^r$, define $\sigma_v: \rset^r \to \rset^r$ as 
\[\sigma_v(y_1, \ldots, y_r) = (\sigma(y_1 - v_1), \ldots, \sigma(y_r - v_r))^\top.\]
The network architecture $( L, \mathbf  p)$ consists of a positive integer $ L$ (depth) and a width vector $\mathbf p = (p_0, \ldots, p_{L+1}) \in \mathbb{N}^{L+2}$. A neural network with architecture $(L, \mathbf p)$ is any function of the form
\[
f: \rset^{p_0} \to \rset^{p_{L+1}}, \quad f(x) = W_L \sigma_{v_L}W_{L-1} \sigma_{v_{L-1}} \ldots W_1 \sigma_{v_1} W_0 x \, .
\]
Denote by $\mathcal F( L, \mathbf p)$ the space of network functions with given network architecture and network parameters bounded by one, that is,
\[
\mathcal F( L,\mathbf  p): = \{f \text{ of the form }: \max_{0\le j\le L} \|W_j\|_\infty \vee \|v_j\|_\infty \le 1\}. 
\]
We may also model the network sparsity assuming that there are only a few nonzero network parameters. Define
\[
\mathcal F(L, \mathbf p, s) = \{f \in \mathcal F(L, \mathbf p): \sum_{j=0}^L \|W_j\|_0 + \|v_j\|_0 \le s \}. 
\]
\begin{proposition}
\label{coordinate approximation}
Suppose \Cref{assum:Holder} holds. Then, for any $a \in \mathsf{A}$ and any integers $m \geq 1$, $J \geq J_0 =  (\beta+1)^d  \vee (K+1) e^d$, there exists a network
\begin{align}
F_a \in \mathcal F\big( L, \big(d, \,6(d + \lceil\beta\rceil)J, \ldots, 6(d + \lceil\beta\rceil)J, 1\big), \,s\big)
\end{align}
with depth \(L = 8 + (m+5)(1 + \lceil\log_2(d\vee\beta)\rceil)\eqsp,\) and with at most
\(
s \le 141 (d + \beta +1)^{3 + d} J (m+6)
\) nonzero parameters,
such that 
\[
\sup_{s\in[0,1]^d}\big|Q^\star(s,a)-F_a(s)\big| \le (2 K +1) (1 + d^2 + \beta^2) 6^d J 2^{-m} + K 3^\beta J^{-\beta/d}.
\]  
\end{proposition}
\begin{proof}
The proof follows from \cite[Theorem 5]{MR4134774}. 
\end{proof}
\paragraph{Clipped ReLU policy class.} For fixed \(\varepsilon_Q\in(0,1)\)
choose
\[
    J
    :=
    \left\lceil
        3^d2^{d/\beta}\varepsilon_Q^{-d/\beta}
    \right\rceil
    \vee J_0\eqsp, \quad     m
    :=
    \left\lceil
        \log_2
        \left(
            \frac{2(2K+1)(1+d^2+\beta^2)6^dJ}
                 {K\varepsilon_Q}
        \right)
    \right\rceil
    \vee1\eqsp,
\]
and let \((L, \mathbf p,s)\) be as in
\Cref{coordinate approximation}.
Then \(J\geq J_0\), as required in
\Cref{coordinate approximation}, and  for every \(a\in\mathsf A\), there exists
\(F_a^\circ\in\mathcal F(L,\mathbf  p,s)\) such that
\begin{equation}
\label{Q approximation}
    \max_{a\in\mathsf A}
    \sup_{s\in[0,1]^d}
    |Q^\star(s,a)-F_a^\circ(s)|
    \leq
    K\varepsilon_Q.
\end{equation}
Moreover, \(J=O(\varepsilon_Q^{-d/\beta})\),  \(m, L =O(\log(1/\varepsilon_Q))\) and  \(s = O(\varepsilon_Q^{-d/\beta} \log (1/\varepsilon_Q)) \),
where the constants may depend on \(d,\beta\), and \(K\). Let
\[
 Q_{\mathrm{NN}}
    :=
    \left\{
        F=(F_a)_{a\in\mathsf A}:
        F_a\in\mathcal F(L,\mathbf p,s)
    \right\}
\]
be the corresponding class of vector-valued logit functions.
For \(q\in\mathbb R^{|\mathsf A|}\), define the statewise
normalization-and-clipping map
\[
    \mathcal C_\Gamma(q)
    :=
    \operatorname{clip}
    \left(
        q-\min_{a\in\mathsf A}q(a)\mathbf 1,\,
        0,\Gamma
    \right),
\]
where the clipping is applied componentwise. For \(F\in Q_{\mathrm{NN}}\),
\begin{align}
\label{eq:clipping}
    \widetilde Q_F(s)
    :=
    \mathcal C_\Gamma(F(s)),
    \qquad
    \pi_F(\cdot\mid s)
    :=
    \operatorname{softmax}
    \left(
         \widetilde Q_F(s)/\lambda
    \right)\eqsp,
\end{align}
and define the policy class \(\mathcal F_{\mathrm{NN}}
    :=
    \left\{
        \pi_F:
        F\in Q_{\mathrm{NN}}
    \right\}.\)
By construction, \(0
    \leq
    \widetilde Q_F(s,a)
    \leq
    \Gamma\)
for every \(F\in Q_{\mathrm{NN}}\) and every
\((s,a)\in[0,1]^d\times\mathsf A\). Hence
\begin{align}
\label{eq:floor}
    \pi_F(a\mid s)
    \geq
    \frac{1}
    {1+(|\mathsf A|-1)e^{\Gamma/\lambda}}
    =\alpha(\Gamma,\lambda,|\mathsf A|).
\end{align}
Thus the entire class \(\mathcal F_{\mathrm{NN}}\) satisfies the uniform-floor
condition in \Cref{assum: F}.
\paragraph{Approximation of the expert policy.} Let
\(
    q^\star(s)
    :=
    \bigl(
        Q^\star(s,a)
    \bigr)_{a\in\mathsf A}.
\)
Under \eqref{eq:spanqstar},
\begin{equation}
\label{eq:clipping-preserves-expert-policy}
    \mathcal C_\Gamma(q^\star(s))
    =
    q^\star(s)
    -
    \min_{a\in\mathsf A}Q^\star(s,a)\,\mathbf 1.
\end{equation}
Since softmax is invariant under the addition of a common constant to
all logits,
\[
    \operatorname{softmax}
    \left(
        {\mathcal C_\Gamma(q^\star(s))}/{\lambda}
    \right)
    =
    \pi^\star(\cdot\mid s).
\]
Let
\(
    F^\circ
    :=
    \bigl(
        F_a^\circ
    \bigr)_{a\in\mathsf A}
    \in
     Q_{\mathrm{NN}}
\)
be the vector of networks defined by
\eqref{Q approximation}. Then,
\begin{equation}
\label{eq:clipped-logit-approximation}
    \left\|
        \mathcal C_\Gamma(F^\circ)
        -
        \mathcal C_\Gamma(q^\star)
    \right\|_\infty
    \leq
    2K\varepsilon_Q.
\end{equation}
Applying \Cref{lem:softmax-KL} statewise and using
\eqref{eq:clipping-preserves-expert-policy} gives
\begin{align}
    \mathcal K(\pi_{F^\circ})
    &=
    \E_{S\sim d_{\pi^\star}}
    \KL\left(
        \operatorname{softmax}
        \left(
            \frac{\mathcal C_\Gamma(q^\star(S))}{\lambda}
        \right)
        \,\middle|\,
        \operatorname{softmax}
        \left(
            \frac{\mathcal C_\Gamma(F^\circ(S))}{\lambda}
        \right)
    \right)
    \nonumber\\
    &\leq
    \frac{1}{2\lambda^2}
    \left\|
        \mathcal C_\Gamma(F^\circ)
        -
        \mathcal C_\Gamma(q^\star)
    \right\|_\infty^2
    \leq
    \frac{2K^2\varepsilon_Q^2}{\lambda^2}.
\label{E est}
\end{align}
Consequently, since \(\mathcal R\equiv0\) in this subsection,
\begin{equation}
\label{eq:ReLU-approximation-error}
    \Delta(\mathcal F_{\mathrm{NN}})
    =
    \inf_{\pi\in\mathcal F_{\mathrm{NN}}}
    \mathcal K(\pi)
    \leq
    \frac{2K^2\varepsilon_Q^2}{\lambda^2}.
\end{equation}

\paragraph{Metric entropy and compactness of \(\mathcal{F}_{\operatorname{NN}}\).}
By \cite[Lemma~5]{MR4134774}, the vector-valued logit class satisfies,
for every \(\eta\in(0,1]\),
\begin{equation}
\label{eq:vector-ReLU-covering}
    \log
    \mathcal N
    \bigl(
        \eta, Q_{\mathrm{NN}},\|\cdot\|_\infty
    \bigr)
    \leq
    |\mathsf A|(s+1)
    \log
    (
        {
            2^{2 L+5}(L+1)d^2 s^{2L}
        }{\eta^{-1}}
    ).
\end{equation}
Since the softmax map is \(1/(2\lambda)\)-Lipschitz, we obtain 
\begin{equation}
\label{eq:network-to-policy-Lipschitz}
    \|\pi_F-\pi_G\|_\infty
    \leq
    \lambda^{-1}
    \|F-G\|_\infty,
    \qquad
    F,G\in Q_{\mathrm{NN}}.
\end{equation}
Set
\(
    D_{\mathrm{NN}}
    :=
    |\mathsf A|(s+1),\)
\(C_{\mathrm{NN}}
    :=
    2^{2L+5}(L+1)d^2 s^{2L},
\)
and
\(
    \Lambda_{\mathrm{NN}}
    :=
    C_{\mathrm{NN}}
    \left(
        1+\lambda^{-1}
    \right).
\)
For \(\varepsilon\in(0,1]\), take
\(\eta=\min\{\lambda\varepsilon,1\}\) in
\eqref{eq:vector-ReLU-covering}. By
\eqref{eq:network-to-policy-Lipschitz},
\[
\begin{aligned}
    \log
    \mathcal N
    \bigl(
        \varepsilon,
        \mathcal F_{\mathrm{NN}},
        \|\cdot\|_\infty
    \bigr)
    &\leq
    D_{\mathrm{NN}}
    \log
    \left(
        \frac{C_{\mathrm{NN}}}{\eta}
    \right)
    \leq
    D_{\mathrm{NN}}
    \log
    \left(
        \frac{\Lambda_{\mathrm{NN}}}{\varepsilon}
    \right).
\end{aligned}
\]
For \(\varepsilon\in(1,\Lambda_{\mathrm{NN}}]\), the same inequality is
immediate because the diameter of a policy class in the uniform metric
is at most one. Hence
\begin{equation}
\label{eq:clipped-policy-covering}
    \log
    \mathcal N
    \bigl(
        \varepsilon,
        \mathcal F_{\mathrm{NN}},
        \|\cdot\|_\infty
    \bigr)
    \leq
    D_{\mathrm{NN}}
    \log
    \left(
        \frac{\Lambda_{\mathrm{NN}}}{\varepsilon}
    \right),
    \qquad
    \varepsilon\in(0,\Lambda_{\mathrm{NN}}].
\end{equation}
 Thus 
\eqref{eq:floor} and
\eqref{eq:clipped-policy-covering} show that
\Cref{assum: F} holds for \(\mathcal F_{\mathrm{NN}}\).
Moreover,
\[
    D_{\mathrm{NN}}
    =
    O\left(
        |\mathsf A|
        \varepsilon_Q^{-d/\beta}
        \log(1/\varepsilon_Q)
    \right)\eqsp, \quad  \log\Lambda_{\mathrm{NN}}
    =
    O\left(
        \log^2(1/\varepsilon_Q)
        +
        \log(1+\lambda^{-1})
    \right),
\]
where the implicit constants may depend on \(d,\beta\), and \(K\).

Let \(\Theta_{L,\mathbf p,s}\) be the parameter set of the scalar
network class. Since all parameters are bounded and the sparsity
constraint is a finite union of closed coordinate subspaces,
\(\Theta_{L,\mathbf p,s}\) is compact. Hence
\(\Theta_{L,\mathbf p,s}^{|\mathsf A|}\) is compact. The maps \eqref{eq:clipping}
are continuous, therefore
\(\mathcal F_{\mathrm{NN}}\) is compact.
\subsection{Final rate}
\label{subsec:relu-final-rate}

Throughout this subsection, the problem parameters
\(d,\beta,\Gamma,|\mathsf A|\), and \(\lambda\) are fixed.
For each \(\varepsilon_Q\in(0,1)\), let
\(\mathcal F_{\mathrm{NN}}(\varepsilon_Q)\) be the clipped ReLU policy
class constructed above. By \eqref{eq:ReLU-approximation-error},
\begin{equation}
\label{eq:relu-Delta-final}
    \Delta\bigl(\mathcal F_{\mathrm{NN}}(\varepsilon_Q)\bigr)
    \lesssim
    \frac{K^2\varepsilon_Q^2}{\lambda^2}.
\end{equation}
Moreover, the network complexity estimates
\eqref{eq:clipped-policy-covering} imply
\[
    D_{\varepsilon_Q}
    \lesssim
    |\mathsf A|\,
    \varepsilon_Q^{-d/\beta}
    \log\frac{e}{\varepsilon_Q},
    \qquad
    \log\Lambda_{\varepsilon_Q}
    \lesssim
    \log^2\frac{e}{\varepsilon_Q},
\]
where the implicit constants may depend on the fixed problem
parameters.
Applying \Cref{thm:hp-reward} to
\(\mathcal F_{\mathrm{NN}}(\varepsilon_Q)\), we obtain, for some
constant \(c>0\), with probability at least \(1-\delta\),
\begin{equation}
\label{eq:relu-risk-before-optimization}
\begin{aligned}
    \mathcal K(\widehat\pi)
    \lesssim\;&
    \frac{K^2\varepsilon_Q^2}{\lambda^2}
    +
    \frac{
        |\mathsf A|\tau_{\mathrm{mix}}
    }{N}
    \varepsilon_Q^{-d/\beta}
    \log^c\left(
        \frac{eN}{\varepsilon_Q}
    \right)
    +
    \frac{\tau_{\mathrm{mix}}}{N}
    \log\left(
        \frac{\sqrt{2\mathfrak N_0}}{\delta}
    \right).
\end{aligned}
\end{equation}
Balancing the polynomial parts of the first two terms gives
\begin{equation}
\label{eq:relu-optimal-epsilon}
    \varepsilon_Q^\star
    :=
    \left(
        \frac{
            |\mathsf A|\lambda^2\tau_{\mathrm{mix}}
        }{
            K^2N
        }
    \right)^{\frac{\beta}{d+2\beta}}.
\end{equation}
For \(N\) sufficiently large, \(\varepsilon_Q^\star\in(0,1)\), and
\(\log(eN/\varepsilon_Q^\star)\lesssim\log N\). Substituting
\eqref{eq:relu-optimal-epsilon} into
\eqref{eq:relu-risk-before-optimization} yields
\begin{equation}
\label{eq:relu-final-KL-rate}
\begin{aligned}
    \mathcal K(\widehat\pi)
    \lesssim\;&
    K^{\frac{2d}{d+2\beta}}
    \lambda^{-\frac{2d}{d+2\beta}}
    \left(
        \frac{
            |\mathsf A|\tau_{\mathrm{mix}}
        }{N}
    \right)^{\frac{2\beta}{d+2\beta}}
    \log^c N
    +
    \frac{\tau_{\mathrm{mix}}}{N}
    \log\left(
        \frac{\sqrt{2\mathfrak N_0}}{\delta}
    \right).
\end{aligned}
\end{equation}

Finally, by the policy-to-reward estimate in
\Cref{thm:hp-reward},
\begin{equation}
\label{eq:relu-final-reward-rate}
\begin{aligned}
    \|\widehat R-R_{\mathrm{LS}}^\star\|_{\rho^\star}^2
    \lesssim\;&
    K^{\frac{2d}{d+2\beta}}
    \left(
        \frac{
            \lambda^2|\mathsf A|\tau_{\mathrm{mix}}
        }{N}
    \right)^{\frac{2\beta}{d+2\beta}}
    \log^c N
    +
    \frac{
        \lambda^2\tau_{\mathrm{mix}}
    }{N}
    \log\left(
        \frac{\sqrt{2\mathfrak N_0}}{\delta}
    \right).
\end{aligned}
\end{equation}
This proves \Cref{thm:holder-relu-rate}.

\subsection{Quadratic KL control for softmax policies}
\begin{lemma}
\label{lem:softmax-KL}
Let \(q,q'\in\mathbb R^{|\mathsf A|}\), and define
\(
    \pi_q = \operatorname{softmax}(q/\lambda).
\)
Then
\[
    \KL(\pi_q\mid\pi_{q'})
    \leq
    \frac{\|q-q'\|_\infty^2}{2\lambda^2}.
\]
\end{lemma}
\begin{proof}
Let \(H(a):=q'_a-q_a\). Since
\[
    \frac{\sum_a\exp(q'_a/\lambda)}
         {\sum_a\exp(q_a/\lambda)}
    =
    \mathbb E_{A\sim\pi_q}
    \exp\left(\frac{H(A)}{\lambda}\right),
\]
we have
\begin{align}
\label{eq:KL_hoeffding}
    \KL(\pi_q\mid\pi_{q'})
    =
    \log
    \mathbb E_{\pi_q}
    \exp\left(\frac{H}{\lambda}\right)
    -
    \frac{\mathbb E_{\pi_q}[H]}{\lambda} = \log
    \mathbb E_{\pi_q}
    \exp\left(\frac{H-  \E_{\pi_q}[H]}{\lambda}\right)
    .
\end{align}
Hoeffding's lemma gives
\[
    \KL(\pi_q\mid\pi_{q'})
    \leq
    \frac{(\max_aH(a)-\min_aH(a))^2}{8\lambda^2}\le \frac{\|q-q'\|_\infty^2}{2\lambda^2}.
\]
\end{proof}

\section{Proof of Theorem \ref{thm:pointwise-lower-bound}}
\label{app:proof-lower-bound}

\begin{proof}
Set \(A:=|\mathsf A|\). We construct a finite family of centered logit functions and apply an
Assouad-type argument. We first assume that \(N\gtrsim N_0\), where 
\begin{equation}
\label{eq:lower-bound-sample-size}
    N_0
   =
     A
    \left[
        \left(\frac{K}{\lambda}\right)^{d/\beta}
        \vee
        \left(\frac{\lambda}{K}\right)^2
    \right].
\end{equation}
Let
\(\varphi\in\mathcal C^\infty_c((-1/4,1/4)^d)\) be non-negative,
non-zero, and  \(\|\varphi\|_{\mathcal C^\beta(\mathbb R^d)}
    \leq1.\)
For every sufficiently small \(h>0\), one can choose points
\(x_1,\ldots,x_m\in[0,1]^d\) such that \(x_k+h[-1/4,1/4]^d\)
are mutually disjoint, contained in \([0,1]^d\), and \(m\geq c_d h^{-d}.\)
Set
\begin{equation}
\label{eq:lower-bound-bumps}
    \psi_k(s)
    :=
    c_\varphi Kh^\beta
    \varphi\left(\frac{s-x_k}{h}\right),
    \qquad
    k=1,\ldots,m,
\end{equation}
where \(c_\varphi>0\) is sufficiently small. By the usual scaling
properties of H\"older norms and the separation of the supports, every
sum of a subcollection of the functions \(\psi_k\) has
\(\mathcal C^\beta\)-norm at most \(K\). Let \(J_A:=\lfloor A/2\rfloor,\)
and choose \(J_A\) disjoint pairs of actions
\[
    (a_j^+,a_j^-),
    \qquad
    j=1,\ldots,J_A.
\]
For \(\theta
    =
    (\theta_{jk})
    \in
    \{0,1\}^{J_A m},\)
set
\begin{equation}
\label{eq:Assouad-logit-family}
    g_\theta(s,a)
    :=
    \sum_{j=1}^{J_A}
    \sum_{k=1}^m
    \theta_{jk}\psi_k(s)v_j(a)\eqsp, \quad v_j(a)
    :=
    \mathbf 1_{\{a=a_j^+\}}
    -
    \mathbf 1_{\{a=a_j^-\}}\eqsp.
\end{equation}
Since the action pairs and the spatial supports are disjoint,
\(g_\theta\in\mathcal G^\beta(K)\) for every \(\theta\).
Choose
\begin{equation}
\label{eq:lower-bound-bandwidth}
    h
    :=
    c_h
    \left(
        \frac{A\lambda^2}{NK^2}
    \right)^{\frac{1}{2\beta+d}},
\end{equation}
where \(c_h>0\) is sufficiently small. Condition
\eqref{eq:lower-bound-sample-size} ensures that \(h\) is small enough
for the packing above and that
\begin{equation}
\label{eq:small-logits-lower-bound}
    \|g_\theta\|_\infty
    \leq
    c_\varphi Kh^\beta
    \leq
    \frac{\lambda}{2}.
\end{equation}

We next bound the \(\KL\) divergence between the corresponding Markov trajectory laws.
Let \(\theta^{(jk)}\) be obtained from \(\theta\) by flipping the
coordinate \((j,k)\). Since the physical transition kernel and the
initial state distribution are the same under both models, the
likelihood ratio of the two path laws contains only the policy ratios.
Consequently,
\begin{align}
\label{eq:pathwise-KL-chain-rule}
    \KL\left(
        \mathbb P_\theta^{(N)}
        \,\middle|\,
        \mathbb P_{\theta^{(jk)}}^{(N)}
    \right)
    &=
    \sum_{t=0}^{N-1}
    \mathbb E_\theta
    \KL\left(
        \pi_{g_\theta}(\cdot\mid S_t)
        \,\middle|\,
        \pi_{g_{\theta^{(jk)}}}(\cdot\mid S_t)
    \right).
\end{align}
Let \(q_{\theta,t}\) denote the density of \(S_t\). By
\Cref{assum:lower-bound-dynamics},
\[
    \|q_{\theta,0}\|_\infty\leq\overline q_0,
    \qquad
    \|q_{\theta,t}\|_\infty\leq\overline p,
    \quad t\geq1.
\]
Indeed, for \(t\geq1\),
\[
\begin{aligned}
    q_{\theta,t}(s')
    &=
    \int_{\mathsf S}
    q_{\theta,t-1}(s)
    \sum_a
    \pi_{g_\theta}(a\mid s)
    p(s'\mid s,a)\,ds
   \leq
    \overline p.
\end{aligned}
\]
Set \(\overline q
    :=
    \overline q_0\vee\overline p.\)
Using \Cref{eq:kl_softmax_e} and the fact that
neighboring parameters differ by
\(\psi_k(s)v_j\), we obtain
\begin{align}
\label{eq:pathwise-neighbor-KL}
    \KL\left(
        \mathbb P_\theta^{(N)}
        \,\middle|\,
        \mathbb P_{\theta^{(jk)}}^{(N)}
    \right)
    &\leq
    \frac{e}{A\lambda^2}
    \sum_{t=0}^{N-1}
    \mathbb E_\theta[\psi_k(S_t)^2]
    \leq
    \frac{
        e\overline qN
    }{
        A\lambda^2
    }
    \int_{\mathsf S}\psi_k(s)^2\,ds
    \leq
    C
    \frac{
        NK^2h^{2\beta+d}
    }{
        A\lambda^2
    }.
    \nonumber
\end{align}
By the choice of \(h\) in \eqref{eq:lower-bound-bandwidth}, the last
quantity is at most \(1/8\), provided \(c_h\) is sufficiently small.
Pinsker's inequality therefore yields
\begin{equation}
\label{eq:neighbor-TV}
    \left\|
        \mathbb P_\theta^{(N)}
        -
        \mathbb P_{\theta^{(jk)}}^{(N)}
    \right\|_{\mathrm{TV}}
    \leq
    \frac14.
\end{equation}

We therefore reduce the reward estimation problem to that of estimating the centered action logits. Let \(\nu(ds,da)
    :=
    \frac1A\,ds\)
denote the product of Lebesgue measure and the uniform distribution on
the action space. Define the action-centering operator
\[
    (\mathcal C_0 u)(s,a)
    :=
    u(s,a) 
    -
    \frac1A
    \sum_{b\in\mathsf A}u(s,b).
\]
This is the orthogonal projector onto the centered action subspace in
\(L_2(\nu)\). For every centered \(g\), \(    \mathcal C_0
    \bigl(
        \lambda\log\pi_g
    \bigr)
    =
    g.\)
Given any reward estimator \(\widehat R_N\), define \(\widehat g_N
    :=
    \mathcal C_0
    \bigl(
        \widehat R_N-\mathcal M f_0
    \bigr).\)
Since \(\mathcal C_0\) is an orthogonal projector,
\begin{equation}
\label{eq:reward-to-centered-logit}
    \|
        \widehat R_N-R_{g_\theta,f_0}
    \|_\nu^2
    \geq
    \|
        \widehat g_N-g_\theta
    \|_\nu^2.
\end{equation}
For every member of the hard family,
\eqref{eq:small-logits-lower-bound} gives \(\pi_{g_\theta}(a\mid s)
    \geq
    {(eA)^{-1}}.\)
Furthermore, the invariant density satisfies
\(q_{g_\theta}\geq\underline p\). Hence, for every measurable \(u\),
\begin{equation}
\label{eq:rho-dominates-nu}
    \|u\|_{\rho_{g_\theta}}^2
    \geq
    e^{-1}\underline p\,
    \|u\|_\nu^2.
\end{equation}
The functions \( b_{jk}(s,a):=\psi_k(s)v_j(a)\)
are mutually orthogonal in \(L_2(\nu)\), and
\begin{equation}
\label{eq:bit-separation}
    \|b_{jk}\|_\nu^2
    =
    \frac{2}{A}
    \int_{\mathsf S}\psi_k(s)^2\,ds
    \geq
    c
    \frac{
        K^2h^{2\beta+d}
    }{A}
    =:\delta_h^2.
\end{equation}
For an arbitrary estimator \(\widehat g_N\), define
\[
    \widehat\theta_{jk}
    :=
    \mathbf 1
    \left\{
        \langle\widehat g_N,b_{jk}\rangle_\nu
        \geq
        \frac12\|b_{jk}\|_\nu^2
    \right\}.
\]
Orthogonality implies
\[
    \|
        \widehat g_N-g_\theta
    \|_\nu^2
    \geq
    \frac{\delta_h^2}{4}
    \sum_{j=1}^{J_A}
    \sum_{k=1}^m
    \mathbf 1_{\{
        \widehat\theta_{jk}\neq\theta_{jk}
    \}}.
\]
Averaging over \(\theta\in\{0,1\}^{J_A m}\), pairing each
\(\theta\) with \(\theta^{(jk)}\), and using
\eqref{eq:neighbor-TV}, we obtain
\begin{align}
    \sup_\theta
    \mathbb E_\theta
    \|
        \widehat g_N-g_\theta
    \|_\nu^2
    &\geq
    \frac{
        J_A m\,\delta_h^2
    }{8}
    \left(
        1-
        \max_{\theta,j,k}
        \|
            \mathbb P_\theta^{(N)}
            -
            \mathbb P_{\theta^{(jk)}}^{(N)}
        \|_{\mathrm{TV}}
    \right) \geq
    cJ_A m\delta_h^2.
\label{eq:Assouad-conclusion}
\end{align}
Since \(J_A\geq A/3\) 
and \(m\geq c_dh^{-d},\)
equation \eqref{eq:bit-separation} yields \( J_A m\delta_h^2
    \geq
    cK^2h^{2\beta}.\)
Combining this bound with
\eqref{eq:reward-to-centered-logit},
\eqref{eq:rho-dominates-nu}, and
\eqref{eq:lower-bound-bandwidth}, we conclude that
\[
\begin{aligned}
    \inf_{\widehat R_N}
    \sup_{g\in\mathcal G^\beta(K)}
    \mathbb E_g
    \|
        \widehat R_N-R_{g,f_0}
    \|_{\rho_g}^2
    &\geq
    cK^2h^{2\beta}=
    c
    K^{\frac{2d}{2\beta+d}}
    \left(
        \frac{\lambda^2A}{N}
    \right)^{\frac{2\beta}{2\beta+d}}.
\end{aligned}
\]
\end{proof}

\begin{lemma}
\label{eq:kl_softmax_e}
    Let \(q,q'\in\mathbb R^{|\mathsf A|}\)  satisfy
\( q'-q
    =
    u(\mathbf e_i- \mathbf e_j),
    \) with \(
    |u|\leq\lambda,\)
and suppose that \(\|q\|_\infty\leq\lambda/2\). Define
\(
    \pi_q = \operatorname{softmax}(q/\lambda).
\)
Then
\begin{equation}
\label{eq:local-softmax-KL-lower-proof}
    \KL(\pi_q\mid\pi_{q'})
    \leq
    \frac{e^2u^2}{A\lambda^2}.
\end{equation}
\end{lemma}
\begin{proof}
Writing \(H=q'-q\), we have as in \eqref{eq:KL_hoeffding}
\[
    \KL(\pi_q\mid\pi_{q'})
    =
    \log\E_{\pi_q}e^{H/\lambda}
    -
    \E_{\pi_q}\big[H/{\lambda}\big].
\]
Since \(\log x\leq x-1\) and
\(e^x-1-x\leq(e/2)x^2\) for \(|x|\leq1\),
\[
\begin{aligned}
    \KL(\pi_q\mid\pi_{q'})
    &\leq
    \E_{\pi_q}
    \left[
        e^{H/\lambda}
        -
        1
        -
        \frac{H}{\lambda}
    \right]
    \leq
    \frac{e}{2\lambda^2}
    \E_{\pi_q}[H^2].
\end{aligned}
\]
Moreover, 
\(    \E_{\pi_q}[H^2]
    =
    u^2\bigl(\pi_q(i)+\pi_q(j)\bigr).\)
Since \(\|q\|_\infty\leq\lambda/2\), \(\pi_q(a)\leq\frac{e}{A}\), we get \eqref{eq:local-softmax-KL-lower-proof}.
\end{proof}

\section{Technical proofs for Section \ref{sec:estimated-bellman}}
\label{app:unknown-operator-proofs}

\subsection{Proof of Lemma \ref{lem:Galerkin-approximation-rate}}
\label{app:proof of galerkin rate}
\begin{proof}
We use the decomposition
\begin{align}
\label{eq:Galerkin-error-decomposition}
    \MKQ-\MKQ_{\dsieve}
    &=
    (\operatorname{Id}_{\mathbb G}
      -\Pi_{\mathcal W_{\dsieve}})\MKQ
   +
    \Pi_{\mathcal W_{\dsieve}}
    \MKQ
    (\operatorname{Id}_{\mathbb H}
      -\Pi_{\mathcal V_{\dsieve}}).
\end{align}
First, for \(f\in\mathbb H\) and \(a\in\mathsf{A}\) we use the shorthand $(\MKQ f)_a(s) = \MKQ f(s, a)$.  Minkowski's and Cauchy--Schwarz
inequalities give
\begin{align}
\label{eq:Pf_smoothness}
    \|(\MKQ f)_a\|_{\mathcal C^\beta_{}(\mathsf S)}^2
    &= 
    \left\|\int_{\mathsf{S}} f(s')\kappa_a(\cdot, s')d_{\pi^\star}(ds')\right\|_{\mathcal C^\beta(\mathsf S)}^2\\
    &\le
   \left( \int_{\mathsf{S}} |f(s')|\left\|\kappa_a(\cdot, s')\right\|_{\mathcal C^\beta(\mathsf S)} d_{\pi^\star}(ds')\right)^2
    \\
    &\le \Big(\int_{\mathsf{S}} |f(s')|^2d_{\pi^\star}(ds')\Big)
\int_{\mathsf{S}} \left\|\kappa_a(\cdot, s')\right\|_{\mathcal C^\beta(\mathsf S)}^2 d_{\pi^\star}(ds') \le 
    L_{\mathrm{\kappa}}^2\|f\|_{\mathbb H}^2,
\end{align}
where the last step follows from \Cref{assum:kernel_smoothness}. For each \(a\in\mathsf A\), by \eqref{eq:Jackson-Vk}, choose
\(v_{\dsieve,a}\in\mathcal V_\dsieve\) such that
\[
    \|(\MKQ f)_a-v_{\dsieve,a}\|_\infty
    \leq
    C_{\mathcal V}\dsieve^{-\beta/d}
    \|(\MKQ f)_a\|_{\mathcal C^\beta(\mathsf S)},
\]
and define
\(\tilde v(s,a):=v_{\dsieve,a}(s)\).
By construction, \(\widetilde v\in\mathcal W_\dsieve\). Hence, 
\begin{align}
\label{eq:best_approx}
    \|(\operatorname{Id}_{\mathbb{G}} - \Pi_{\mathcal{W}_\dsieve})\MKQ f\|_{\mathbb{G}} = \inf_{v\in \mathcal{W}_\dsieve}\|\MKQ f - v\|_{\mathbb{G}} \le \|\MKQ f - \tilde v\|_{\mathbb{G}}, 
\end{align}
Therefore, by \eqref{eq:Pf_smoothness}
\begin{align}
\label{eq:v_tilde_estimation}
    \|\MKQ f - \tilde v\|_{\mathbb{G}}^2 &= \int_{\mathsf{S}\times\mathsf{A}} (\tilde v_a(s) - (\MKQ f)_a(s))^2\rho^\star(ds, da)\\
    &\le \int_{\mathsf{S}\times\mathsf{A}} \|\tilde v_a - (\MKQ f)_a\|_{\infty}^2\,\rho^\star(ds, da)\\
    &\le C_{\mathcal V}^2
        \dsieve^{-2\beta/d} \int_{\mathsf{S}\times\mathsf{A}}\|(\MKQ f)_a\|_{\mathcal{C}^{\beta}(\mathsf{S})}^2\,\rho^\star(ds, da) \le  C_{\mathcal V}^2
        \dsieve^{-2\beta/d}
L_{\mathrm{\kappa}}^2\| f\|_{\mathbb H}^2,
\end{align}
Finally, \eqref{eq:v_tilde_estimation} and \eqref{eq:best_approx} yield 
\begin{align}
\label{eq:Galerkin-output-term}
    \|(\operatorname{Id}_{\mathbb G}
      -\Pi_{\mathcal W_{\dsieve}})\MKQ\|_{\mathbb H\to\mathbb G}
      &= \sup_{\|f\|_{\mathbb H} = 1} \|(\operatorname{Id}_{\mathbb{G}} - \Pi_{\mathcal{W}_\dsieve})\MKQ f\|_{\mathbb{G}} \le L_{{\kappa}} C_{\mathcal V}
        \dsieve^{-\beta/d}\eqsp.
\end{align}
The adjoint of \(\MKQ\) is given by
\[
    (\MKQ^\ast g)(s')
    =
    \int_{\mathsf S\times\mathsf A}
    \kappa_a(s,s')g(s,a)\,\rho^\star(ds,da).
\]
Hence,
\begin{align}
    \|\MKQ^\ast g\|^2_{\mathcal C^\beta(\mathsf S)}& = \left\|\int_{\mathsf S\times\mathsf A}
\kappa_a(s,\cdot)g(s,a)\,\rho^\star(ds,da)\right\|_{\mathcal C^\beta(\mathsf S)}^2
\\
&\le \left(\int_{\mathsf S\times\mathsf A}
\|\kappa_a(s,\cdot)\|_{\mathcal C^\beta(\mathsf S)}|g(s,a)|\,\rho^\star(ds,da)\right)^2\\
&\le \int_{\mathsf S\times\mathsf A}
|g(s,a)|^2\,\rho^\star(ds,da) \int_{\mathsf S\times\mathsf A}
\|\kappa_a(s,\cdot)\|_{\mathcal C^\beta(\mathsf S)}^2\,\rho^\star(ds,da)
\le L_{\kappa}^2\|g\|_{\mathbb G}^2
\end{align}
Since \(\Pi_{\mathcal V_{\dsieve}}\) is self-adjoint,
\begin{align}
    \|\MKQ
      (\operatorname{Id}_{\mathbb H}
       -\Pi_{\mathcal V_{\dsieve}})\|_{\mathbb H\to\mathbb G}
    &=
    \|(\operatorname{Id}_{\mathbb H}
       -\Pi_{\mathcal V_{\dsieve}})
      \MKQ^\ast\|_{\mathbb G\to\mathbb H}
    \leq
    L_{\kappa} C_{\mathcal V}
    \dsieve^{-\beta/d}.
\end{align}
Combining this estimate with
\eqref{eq:Galerkin-output-term},
\eqref{eq:Galerkin-error-decomposition}, and
\(\|\Pi_{\mathcal W_{\dsieve}}\|=1\) proves the claim.
\end{proof}

\subsection{Proof of Lemma \ref{lem:range-projector}}
\label{app:proof-range-projector}
\begin{proof}
For \(y=Af\in\operatorname{Ran}(A)\),
\[
    \operatorname{dist}
    \bigl(y,\operatorname{Ran}(B)\bigr)
    \leq
    \|(A-B)f\|
    \leq
    \frac{\|A-B\|}{\sigma_A}\|y\|.
\]
Hence
\[
    \|(\operatorname{Id}_{\mathbb G}-\Pi_B)\Pi_A\|
    \leq
    \frac{\|A-B\|}{\sigma_A}.
\]
The analogous bound with \(A\) and \(B\) interchanged also holds. Finally,
\[
    \Pi_A-\Pi_B
    =
    (\operatorname{Id}_{\mathbb G}-\Pi_B)\Pi_A
    -
    \Pi_B(\operatorname{Id}_{\mathbb G}-\Pi_A),
\]
and \eqref{eq:range-projector-perturbation} follows by the triangle inequality.
\end{proof}

\subsection{Proof of Lemma \ref{lem:theta_equation}}
\label{app:theta_equation}
\begin{proof}
Every \(f\in\mathbb H\) admits a unique decomposition
\[
    f=\Phi_\dsieve\theta+h,
    \qquad
    \theta\in\mathbb R^\dsieve,
    \quad
    h\in\mathcal V_\dsieve^\perp.
\]
Since \(\Pi_{\mathcal V_\dsieve}h=0\), the definition of \(\mathcal M_\dsieve\) yields
\begin{equation}
\label{eq:Mk-splitting}
    \mathcal M_\dsieve(\Phi_\dsieve\theta+h)
    =
    \mathcal M_\dsieve\Phi_\dsieve\theta+\mathcal I h.
\end{equation}
Hence
\begin{align}
    f_\dsieve^\star
    &=
    \argmin_{f\in\mathbb H}
    \bigl\|
        \widehat r+\mathcal M_\dsieve f
    \bigr\|_{\mathbb G}^2=\argmin_{\substack{\theta\in\mathbb R^\dsieve\\ h\in\mathcal V_\dsieve^\perp}}
    \bigl\|
        \widehat r+\mathcal M_\dsieve\Phi_\dsieve\theta+\mathcal I h
    \bigr\|_{\mathbb G}^2.
\end{align}
For a fixed \(\theta\in\mathbb R^\dsieve\), let
\[
    h(\theta)
    :=
    \argmin_{h\in\mathcal V_\dsieve^\perp}
    \bigl\|
        \widehat r+\mathcal M_\dsieve\Phi_\dsieve\theta+\mathcal I h
    \bigr\|_{\mathbb G}^2.
\]
The corresponding normal equations imply that, for every
\(v\in\mathcal V_\dsieve^\perp\),
\[
    \left\langle
        \widehat r+\mathcal M_\dsieve\Phi_\dsieve\theta+\mathcal I h(\theta),
        \mathcal I v
    \right\rangle_{\mathbb G}
    =0.
\]
Using \(\mathcal I^\ast\mathcal I=\operatorname{Id}_{\mathbb H}\), this is equivalent to
\[
    \left\langle
        (\operatorname{Id}_{\mathbb H}-\Pi_{\mathcal V_\dsieve})
        \mathcal I^\ast
        \bigl(
            \widehat r+\mathcal M_\dsieve\Phi_\dsieve\theta
        \bigr)
        +h(\theta),
        v
    \right\rangle_{\mathbb H}
    =0
\]
for every \(v\in\mathcal V_\dsieve^\perp\). Therefore,
\begin{equation}
\label{eq:optimal-h-theta}
    h(\theta)
    =
    -
    (\operatorname{Id}_{\mathbb H}-\Pi_{\mathcal V_\dsieve})
    \mathcal I^\ast
    \bigl(
        \widehat r+\mathcal M_\dsieve\Phi_\dsieve\theta
    \bigr).
\end{equation}
Substituting \(h(\theta)\) from \eqref{eq:optimal-h-theta} and using the definition of \(\mathcal Q_\dsieve\) in \eqref{eq:Qk-projector}, we obtain
\begin{equation}
\label{eq:theta-k-minimization}
    \theta_\dsieve^\star
    =
    \arg\min_{\theta\in\mathbb R^\dsieve}
    \left\|
        \mathcal Q_\dsieve
        \bigl(
            \widehat r+\mathcal M_\dsieve\Phi_\dsieve\theta
        \bigr)
    \right\|_{\rho^\star}^{2}.
\end{equation}
Since \(\mathcal Q_\dsieve\) is an orthogonal projector, the normal equations for
\eqref{eq:theta-k-minimization} reduce to
\begin{equation}
\label{eq:normal-equations-operator_app}
    \Phi_\dsieve^\ast\mathcal M_\dsieve^\ast\mathcal Q_\dsieve
    \bigl(
        \widehat r+\mathcal M_\dsieve\Phi_\dsieve\theta_\dsieve^\star
    \bigr)
    =0.
\end{equation}
The solution is unique because, for every \(\theta\in\mathbb R^\dsieve\),
\begin{align}
\label{eq:finite-normal-coercivity}
    \|\mathcal Q_\dsieve\mathcal M_\dsieve\Phi_\dsieve\theta\|_{\rho^\star}^{2}
    &=
    \inf_{h\in\mathcal V_\dsieve^\perp}
    \|\mathcal M_\dsieve(\Phi_\dsieve\theta+h)\|_{\rho^\star}^{2}
    \\
    &\geq
    (1-\gamma)^2
    \inf_{h\in\mathcal V_\dsieve^\perp}
    \|\Phi_\dsieve\theta+h\|_{d_{\pi^{\star}}}^{2}
    \\
    &=
    (1-\gamma)^2
    \|\Phi_\dsieve\theta\|_{d_{\pi^{\star}}}^{2}.
\end{align}
Finally, substituting \(h(\theta_\dsieve^\star)\) from
\eqref{eq:optimal-h-theta} into the definition of \(\widehat R_\dsieve\) yields
\begin{align*}
    \widehat R_\dsieve
   =
    \mathcal Q_\dsieve
    \bigl(
        \widehat r+\mathcal M_\dsieve\Phi_\dsieve\theta_\dsieve^\star
    \bigr),
\end{align*}
which proves \eqref{eq:Rk-Qk-form}.
\end{proof}

\subsection{Equivalence between the normal equation and the coupled system}
\label{app:coupled-system-equivalence}

We first record several useful properties of the feature vectors
\(\Phi_\dsieve\) and \(\Psi_\dsieve\). By the construction in
\eqref{eq:Psi_definition},
\(
    \|\Psi_{\dsieve}(s,a)\|
    =
    \|\Phi_{\dsieve}(s)\|.
\)
Therefore, by \eqref{eq:bounded-features},
\begin{align}
\label{eq:psi_bounded_features}
    \mathfrak B_{\phi,\dsieve}
    =
    \sup_{s\in\mathsf S}
    \|\Phi_{\dsieve}(s)\|
    =
    \sup_{(s,a)\in\mathsf S\times\mathsf A}
    \|\Psi_{\dsieve}(s,a)\|.
\end{align}
Since \(\Phi_\dsieve\) is a basis of \(\mathcal V_\dsieve\), its components are
linearly independent in \(L_2(d_{\pi^\star})\), and hence
\(G_{\phi,\dsieve}\succ0\). Moreover, the definition of
\(\psi_{j,a_0}(s,a)\) in \eqref{eq:Psi_definition}, after a suitable reordering
of the coordinates, implies that \(G_{\psi,\dsieve}\) is block diagonal with
blocks
\[
    G_{\phi,\dsieve}^{(a)}
    =
    \E_{s\sim d_{\pi^\star}}
    \left[
        \pi^\star(a\mid s)
        \Phi_\dsieve(s)\Phi_\dsieve^\top(s)
    \right],
    \qquad a\in\mathsf A.
\]
Since \(\pi^\star\) is a softmax policy,
\(\pi^\star(a\mid s)>0\) \(d_{\pi^\star}\)-a.s. for every
\(a\in\mathsf A\). Together with \(G_{\phi,\dsieve}\succ0\), this implies
\(G_{\phi,\dsieve}^{(a)}\succ0\) for every \(a\in\mathsf A\), and consequently
\(G_{\psi,\dsieve}\succ0\).

Using the explicit representation
\(
    \Pi_{\mathcal{W}_\dsieve}
    =
    \Psi_\dsieve G_{\psi,\dsieve}^{-1}\Psi_\dsieve^\ast
\)
together with \eqref{eq:population-moments-1}, we obtain
\begin{equation}
\label{eq:Mk-Phi}
    \MKQ_\dsieve\Phi_\dsieve
    =
    \Psi_\dsieve G_{\psi,\dsieve}^{-1}B_\dsieve,
    \qquad
    \mathcal M_\dsieve\Phi_\dsieve
    =
    \mathcal I\Phi_\dsieve
    -
    \gamma\Psi_\dsieve G_{\psi,\dsieve}^{-1}B_\dsieve.
\end{equation}
We show that the normal equation \eqref{eq:normal-equations-operator} is equivalent to the coupled system \eqref{eq:coupled-system}. For \(\theta\in\mathbb R^k\), define
\(
    y_\theta
    :=
    \widehat r+\mathcal M_\dsieve\Phi_\dsieve\theta.
\)
Since  \(G_{\phi,\dsieve}\succ0\) and \(G_{\psi,\dsieve}\succ0\), the first two equations in 
\eqref{eq:coupled-system} uniquely determine
\begin{equation}
\label{eq:appendix-u-z}
    u(\theta)
    :=
    G_{\psi,\dsieve}^{-1}B_\dsieve\theta,
    \qquad
    z(\theta)
    :=
    G_{\phi,\dsieve}^{-1}
    \bigl(
        \gamma C_\dsieve^\top u(\theta)-q_\dsieve
    \bigr).
\end{equation}
By \eqref{eq:Mk-Phi},
\begin{equation}
\label{eq:appendix-y-theta}
    y_\theta
    =
    \widehat r
    +
    \mathcal I\Phi_\dsieve\theta
    -
    \gamma\Psi_\dsieve u(\theta).
\end{equation}

We next identify the action of \(\mathcal Q_\dsieve\) on \(y_\theta\). Using
\(\mathcal I^\ast\mathcal I=\operatorname{Id}_{\mathbb H}\) and the definitions of
\(q_\dsieve\) and \(C_\dsieve\), we obtain
\begin{align}
    \Phi_\dsieve^\ast\mathcal I^\ast y_\theta
    &=
    q_\dsieve
    +
    G_{\phi,\dsieve}\theta
    -
    \gamma C_\dsieve^\top u(\theta)
    \nonumber=
    G_{\phi,\dsieve}\bigl(\theta-z(\theta)\bigr).
\label{eq:appendix-projected-Istar-y}
\end{align}
Since
\(
    \Pi_{\mathcal V_\dsieve}
    =
    \Phi_\dsieve G_{\phi,\dsieve}^{-1}\Phi_\dsieve^\ast\eqsp,
\)
it follows that
\begin{equation}
\label{eq:appendix-projected-component}
    \Pi_{\mathcal V_\dsieve}\mathcal I^\ast y_\theta
    =
    \Phi_\dsieve\bigl(\theta-z(\theta)\bigr).
\end{equation}
Moreover, the definition \eqref{eq:Qk-projector} can be written as
\(
    \mathcal Q_\dsieve
    =
    \operatorname{Id}_{\mathbb G}
    -
    \mathcal I\mathcal I^\ast
    +
    \mathcal I\Pi_{\mathcal V_\dsieve}\mathcal I^\ast.
\)
Combining this identity with \eqref{eq:appendix-y-theta} and
\eqref{eq:appendix-projected-component} yields
\begin{equation}
\label{eq:appendix-Qk-y}
    \mathcal Q_\dsieve y_\theta
    =
    (\operatorname{Id}_{\mathbb G}-\mathcal I\mathcal I^\ast)\widehat r
    -
    \gamma
    (\operatorname{Id}_{\mathbb G}-\mathcal I\mathcal I^\ast)
    \Psi_\dsieve u(\theta)
    +
    \mathcal I\Phi_\dsieve\bigl(\theta-z(\theta)\bigr).
\end{equation}
Define
\begin{equation}
\label{eq:appendix-v-definition}
    v(\theta)
    :=
    G_{\psi,\dsieve}^{-1}
    \Psi_\dsieve^\ast\mathcal Q_\dsieve y_\theta.
\end{equation}
Applying \(\Psi_\dsieve^\ast\) to \eqref{eq:appendix-Qk-y} and using the definitions
of \(d_\dsieve\), \(H_\dsieve\), and \(C_\dsieve\), we obtain
\begin{equation}
\label{eq:appendix-v-equation}
    G_{\psi,\dsieve}v(\theta)
    +
    \gamma H_\dsieve u(\theta)
    +
    C_\dsieve z(\theta)
    -
    C_\dsieve\theta
    -
    d_\dsieve
    =
    0.
\end{equation}
Thus, the first three equations in \eqref{eq:coupled-system} are precisely the
finite-dimensional representations of the intermediate quantities
\(u(\theta)\), \(z(\theta)\), and \(v(\theta)\).
It remains to identify the last equation. Since
\(
    \mathcal I^\ast
    (\operatorname{Id}_{\mathbb G}-\mathcal I\mathcal I^\ast)
    =
    0,
\)
equation \eqref{eq:appendix-Qk-y} gives
\begin{align}
    \Phi_\dsieve^\ast\mathcal I^\ast\mathcal Q_\dsieve y_\theta
    &=
    G_{\phi,\dsieve}\bigl(\theta-z(\theta)\bigr)
    =
    G_{\phi,\dsieve}\theta
    -
    \gamma C_\dsieve^\top u(\theta)
    +
    q_\dsieve.
\label{eq:appendix-Istar-Qy}
\end{align}
Combining \eqref{eq:Mk-Phi},
\eqref{eq:appendix-v-definition}, and \eqref{eq:appendix-Istar-Qy}, we obtain
\begin{align}
    \Phi_\dsieve^\ast\mathcal M_\dsieve^\ast\mathcal Q_\dsieve y_\theta
    =
    G_{\phi,\dsieve}\theta
    -
    \gamma C_\dsieve^\top u(\theta)
    -
    \gamma B_\dsieve^\top v(\theta)
    +
    q_\dsieve.
\label{eq:appendix-normal-equation-expanded}
\end{align}
Consequently,
\[
    \Phi_\dsieve^\ast\mathcal M_\dsieve^\ast\mathcal Q_\dsieve
    \bigl(
        \widehat r+\mathcal M_\dsieve\Phi_\dsieve\theta
    \bigr)
    =0
\]
if and only if
\[
    G_{\phi,\dsieve}\theta
    -
    \gamma C_\dsieve^\top u(\theta)
    -
    \gamma B_\dsieve^\top v(\theta)
    +
    q_\dsieve
    =0,
\]
which is exactly the fourth equation in \eqref{eq:coupled-system}.
Hence, if \(\theta_\dsieve^\star\) solves \eqref{eq:normal-equations-operator}, then
\(u_\dsieve^\star:=u(\theta_\dsieve^\star)\),
\(z_\dsieve^\star:=z(\theta_\dsieve^\star)\), and
\(v_\dsieve^\star:=v(\theta_\dsieve^\star)\) satisfy
\eqref{eq:coupled-system}. Conversely, any quadruple
\((\theta_\dsieve^\star,u_\dsieve^\star,z_\dsieve^\star,v_\dsieve^\star)\) satisfying
\eqref{eq:coupled-system} necessarily satisfies
\eqref{eq:appendix-u-z} and \eqref{eq:appendix-v-definition}, and substitution
into the last equation yields \eqref{eq:normal-equations-operator}. This proves
the claimed equivalence.

\subsection{Proof of Proposition \ref{prop:plugin-centering-error}}
\label{app:proof-plugin-centering}
Set \(\mathfrak B_r = \lambda\log\left(\frac{1}{\alpha}\right)\) and
define plug-in errors
\[
    \mathbf{e}_{\psi,\dsieve}(s)
    :=
    \overline\Psi_\dsieve^{\widehat\pi}(s)
    -
    \overline\Psi_\dsieve^\star(s),
    \qquad
    \mathbf{e}_r(s)
    :=
    \overline r^{\,\widehat\pi}(s)
    -
    \overline r^{\,\star}(s).
\]

By Pinsker's inequality and the definition of \(\mathcal K(\widehat\pi)\) in \eqref{eq:kl_pi}, we have
\begin{align}
    &\E_{S\sim d_{\pi^\star}} [\|\mathbf{e}_{\psi,\dsieve}\|^2] 
    = \E_{S\sim d_{\pi^\star}} \left[\left\| \int_{\mathsf A}\Psi_{\dsieve} (s,a) \big(\widehat \pi (da\mid s) -  \pi^\star (da\mid s)\big) \right\|^2\right]\\
    &\qquad\le  4\sup_{s,a}\|\Psi_\dsieve(s,a)\|^2\,\E_{S\sim d_{\pi^\star}} \left[\left\|  \big(\widehat \pi (da\mid s) -  \pi^\star (da\mid s)\big) \right\|_{\operatorname{TV}}^2\right]\\
    &\qquad\le 2  \mathfrak B_{\phi,\dsieve}^2\E_{S\sim d_{\pi^\star}} \left[\KL( \pi^\star (da\mid s)\mid \widehat \pi (da\mid s)) \right] =  2  \mathfrak B_{\phi,\dsieve}^2 \mathcal K(\widehat\pi)
\end{align}
A similar argument for \(\mathbf{e}_r(s)\) yields
\begin{equation}
\label{eq:center-errors-Pinsker}
    \|\mathbf{e}_{\psi,\dsieve}\|_{L_2(d_{\pi^\star})}
    \leq
    \sqrt{2} \mathfrak B_{\phi,\dsieve}\sqrt{\mathcal K(\widehat\pi)},
    \qquad
    \|\mathbf{e}_r\|_{L_2(d_{\pi^\star})}
    \leq
    \sqrt{2}\, \mathfrak B_r\sqrt{\mathcal K(\widehat\pi)}.
\end{equation}
We first control the perturbation of the centered population moments. Observe that \(H_\dsieve\) and \(d_\dsieve\) admit the integral representations
\begin{align}
\label{eq:H-d}
    H_\dsieve
    &:=
    \mathbb E_{\rho^\star}
    \left[
        \big(\Psi_\dsieve-\overline\Psi_\dsieve^{\star}\big)
        \big(\Psi_\dsieve-\overline\Psi_\dsieve^{\star}\big)^\top
    \right],\quad
    d_\dsieve
    :=
    \mathbb E_{\rho^\star}
    \left[
        \big(\Psi_\dsieve-\overline\Psi_\dsieve^{\star}\big)
        \widehat r
    \right].
\end{align}
It follows from \eqref{eq:plugin-H-d} and \eqref{eq:H-d} that
\begin{align}
\label{eq:plugin-H-exact-difference}
    H_\dsieve^{\widehat\pi}-H_\dsieve
    &=
    \mathbb E_{S\sim d_{\pi^\star}}
    \left[
        \mathbf{e}_{\psi,\dsieve}(S)
        \mathbf{e}_{\psi,\dsieve}(S)^\top
    \right]
    \succeq 0,
    \\
\label{eq:plugin-d-exact-difference}
    d_\dsieve^{\widehat\pi}-d_\dsieve
    &=
    -
    \mathbb E_{S\sim d_{\pi^\star}}
    \left[
        \mathbf{e}_{\psi,\dsieve}(S)
        \overline r^{\,\star}(S)
    \right].
\end{align}
Hence, by \eqref{eq:center-errors-Pinsker},
\begin{equation}
\label{eq:plugin-H-d-bounds}
    \|H_\dsieve^{\widehat\pi}-H_\dsieve\|
    \leq
    2\mathfrak {B}_{\phi,\dsieve}^2\mathcal K(\widehat\pi),
    \qquad
    \|d_\dsieve^{\widehat\pi}-d_\dsieve\|
    \leq
    \sqrt{2}\,\mathfrak {B}_{\phi,\dsieve}\mathfrak B_r
    \sqrt{\mathcal K(\widehat\pi)}.
\end{equation}

We next control the induced perturbation of the solution of
\eqref{eq:coupled-system}. Define the Schur complements \(\Delta_\dsieve\) and \(\Delta_\dsieve^{\widehat\pi}\) associated with the systems \eqref{eq:TTSA-population-form} and \eqref{eq:TTSA-population-form-perturb}, respectively,
\begin{align}
    \Delta_\dsieve = A_{11,\dsieve} - A_{12,\dsieve} A_{22, \dsieve}^{-1} A_{21,\dsieve}\eqsp, \qquad \Delta_\dsieve^{\widehat\pi} = A_{11,\dsieve} - A_{12,\dsieve} \big(A_{22, \dsieve}^{\widehat\pi}\big)^{-1} A_{21,\dsieve}\eqsp.
\end{align}
Both Schur complements are invertible. Indeed, by \eqref{eq:finite-normal-coercivity} and \eqref{eq:plugin-H-exact-difference}, 
\begin{equation}
\label{eq:plugin-Schur-stability}
    \Delta_\dsieve^{\widehat\pi}
    \succeq
    \Delta_\dsieve
    \succeq
    (1-\gamma)^2G_{\phi,\dsieve}
    \succ0.
\end{equation}
Eliminating \(u_\dsieve^\star\), \(z_\dsieve^\star\), and
\(v_\dsieve^\star\) from \eqref{eq:coupled-system} yields
\begin{align}
\label{eq:eliminated-slow-equation}
&\Delta_\dsieve\theta_\dsieve^\star
    =
    \ell_\dsieve,
    &\ell_\dsieve
    :=
    \gamma
    B_\dsieve^\top
    G_{\psi,\dsieve}^{-1}
    \bigl(
        d_\dsieve+C_\dsieve G_{\phi,\dsieve}^{-1}q_\dsieve
    \bigr)
    -
    q_\dsieve, \\ 
&\Delta_\dsieve^{\widehat\pi}
\theta_{\dsieve,\widehat\pi}^\star= \ell_\dsieve^{\widehat\pi}, &\ell_\dsieve^{\widehat \pi}
    :=
    \gamma
    B_\dsieve^\top
    G_{\psi,\dsieve}^{-1}
    \bigl(
        d_\dsieve^{\widehat\pi}+C_\dsieve G_{\phi,\dsieve}^{-1}q_\dsieve
    \bigr)
    -
    q_\dsieve.
\end{align}
A direct computation therefore gives
\begin{align}
\label{eq:plugin-Delta-identity}
    \Delta_\dsieve^{\widehat\pi}
    -
    \Delta_\dsieve
    &=
    \gamma^2
    B_\dsieve^\top
    G_{\psi,\dsieve}^{-1}
    \bigl(
        H_\dsieve^{\widehat\pi}-H_\dsieve
    \bigr)
    G_{\psi,\dsieve}^{-1}
    B_\dsieve,
    \\
\label{eq:plugin-ell-identity}
    \ell_\dsieve^{\widehat\pi}
    -
    \ell_\dsieve
    &=
    \gamma
    B_\dsieve^\top
    G_{\psi,\dsieve}^{-1}
    \bigl(
        d_\dsieve^{\widehat\pi}-d_\dsieve
    \bigr).
\end{align}
Subtracting the exact and plug-in slow equations gives
\[
    \theta_{\dsieve,\widehat\pi}^\star-\theta_\dsieve^\star
    =
    (\Delta_\dsieve^{\widehat\pi})^{-1}
    \left[
        \ell_{\dsieve}^{\widehat\pi}-\ell_\dsieve
        -
        (\Delta_\dsieve^{\widehat\pi}-\Delta_\dsieve)
        \theta_\dsieve^\star
    \right].
\]
Combining \eqref{eq:bounded-features} \eqref{eq:plugin-H-d-bounds}--\eqref{eq:plugin-Schur-stability},
 yields
\begin{equation}
\label{eq:plugin-theta-bound}
    \|
        \theta_{\dsieve,\widehat\pi}^\star
        -
        \theta_\dsieve^\star
    \|
    \leq
    \frac{\mathfrak{B}_{\phi,\dsieve}^{11}\mathfrak B_r}{(1-\gamma)^4\lambda_{\min}^3(G_{\phi,\dsieve})\lambda_{\min}^3(G_{\psi,\dsieve})}
    \left(
        \sqrt{\mathcal K(\widehat\pi)}
        +
        \mathcal K(\widehat\pi)
    \right).
\end{equation}
The first two equations of \eqref{eq:coupled-system} give
\begin{align}
\label{eq:plug_in_diff}
u_{\dsieve,\widehat\pi}^\star -  u_\dsieve^\star
    =
    G_{\psi,\dsieve}^{-1}
B_\dsieve(\theta^\star_{\dsieve, \widehat\pi}-\theta_\dsieve^\star)\eqsp,
    \quad
z_{\dsieve,\widehat\pi}^\star - z_\dsieve^\star
    =
    G_{\phi,\dsieve}^{-1}
        \gamma C_\dsieve^\top (u_{\dsieve,\widehat\pi}^\star - u_\dsieve^\star).
\end{align}
Substituting \eqref{eq:plugin-theta-bound} into \eqref{eq:plug_in_diff} yields the corresponding perturbation bounds for the fast variables. It remains to compare the corresponding reward functions. Using
\eqref{eq:Rk-coordinate-representation} and
\eqref{eq:Rk-plugin-target}, we obtain
\begin{align*}
    \widehat R_{\dsieve,\widehat\pi}
    -
    \widehat R_\dsieve
    &=
    -\mathbf{e}_r
    -
    \gamma
    \bigl(
        \Psi_\dsieve
        -
        \overline\Psi_\dsieve^{\widehat\pi}
    \bigr)^\top
    \bigl(
        u_{\dsieve,\widehat\pi}^\star
        -
        u_\dsieve^\star
    \bigr)
    \\
    &\qquad
    +
    \gamma
    \mathbf{e}_{\psi,\dsieve}^\top
    u_\dsieve^\star
    +
    \Phi_\dsieve^\top
    \left[
        \bigl(
            \theta_{\dsieve,\widehat\pi}^\star
            -
            \theta_\dsieve^\star
        \bigr)
        -
        \bigl(
            z_{\dsieve,\widehat\pi}^\star
            -
            z_\dsieve^\star
        \bigr)
    \right].
\end{align*}
Finally, \eqref{eq:center-errors-Pinsker} and 
\eqref{eq:plugin-theta-bound} give
\[
    \|
        \widehat R_{\dsieve,\widehat\pi}
        -
        \widehat R_\dsieve
    \|_{\rho^\star}
    \leq
    \frac{\mathfrak{B}_{\phi,\dsieve}^{16}\lambda\log(1/\alpha)}{(1-\gamma)^4\lambda_{\min}^4(G_{\phi,\dsieve})\lambda_{\min}^4(G_{\psi,\dsieve})}
    \left(
        \sqrt{\mathcal K(\widehat\pi)}
        +
       \mathcal K(\widehat\pi)
    \right).
\]
It remains to invoke the high-probability bound established in \Cref{thm:hp-reward}. In particular, \(\sqrt{\mathcal K(\widehat\pi)}\leq \mathfrak E_N(\delta)\) with probability at least \(1-\delta\), which yields the desired estimate.

\subsection{TTSA Framework}
\label{app:ttsa_framework}
We adopt the notation and results of \cite{butyrin2026gaussian}. We also refer to
\cite{kwon2024two,haque2023tight,doan2021finite,kaledin2020finite} for related finite-time analyses of TTSA with Markovian noise. These works are primarily concerned with mean-square error bounds, whereas, to the best of our knowledge, \cite{butyrin2026gaussian} is the only work that establishes moment bounds of arbitrary order.  We focus on the linear two-timescale SA problem, that is, we aim to find the unique solution $(\theta^\star, w^\star)$ that solves the system of linear equations:
\begin{align} 
\label{eq:linear_sys}
A_{11} \theta + A_{12} w = b_1\eqsp, \quad A_{21} \theta + A_{22} w = b_2\eqsp,
\end{align}
 Assume that the underlying matrices $A_{ij}$ and vectors $b_i$, $i,j \in \{1,2\}$, are not accessible. Instead, we assume that we have access to a sequence of random variables $\{X_k\}_{k \in \nset}$ taking values in a measurable space $(\Xset,\Xsigma)$, and  functions $\funcbw_i(x)$, $\funcAw_{ij}(x)$, $i,j \in \{1,2\}$, which serves as stochastic estimators of $b_i$ and $A_{ij}$, respectively. The corresponding recurrence runs as 
\begin{equation}
\label{eq:tts-gen1}
\begin{split}
\theta_{k+1} &= \theta_k + \beta_k \{ \funcbw^{}_{1}(X_{k+1}) - \funcAw^{}_{11}(X_{k+1}) \theta_k - \funcAw^{}_{12}(X_{k+1}) w_k \}\eqsp, \\
w_{k+1} &= w_k + \gamma_k \{ \funcbw^{}_{2}(X_{k+1}) - \funcAw^{}_{21}(X_{k+1}) \theta_k - \funcAw^{}_{22}(X_{k+1})w_k \}\eqsp.
\end{split}
\end{equation}
For completeness, we state below the assumptions under which the convergence guarantees hold.

\begin{assumption}
\label{assum:hurwitz}
Matrices $-A_{22}$ and $ - \Delta =-\left(A_{11}-A_{12}A_{22}^{-1}A_{21}\right)$ are \textit{Hurwitz}. 
\end{assumption}
\begin{assumption}
\label{assum:aij_bound}
There exist $\bConst{A}, \bConst{b} > 0$ such that for any $i, j \in \{1, 2\}$, 
\begin{align}
\sup\nolimits_{x \in \Xset} \|\funcAw_{ij}(x)\| \leq \bConst{A}\eqsp, \qquad \sup\nolimits_{x \in \Xset} \|\funcbw_{i}(x)\|   \leq \bConst{b} \,.
\end{align}
\end{assumption}
\begin{assumption}
\label{assum:UGE_ttsa}
The sequence $\{X_k\}_{k \in \nset}$ is a Markov chain taking values in a Polish space $(\Xset,\Xsigma)$ with the Markov kernel $\MKQ$. Moreover, $\MKQ$ admits $\pi$ as a unique invariant distribution and is uniformly geometrically ergodic with mixing time $\tmix \in \nset$. In addition, for all $k \in \nset$ and $i, j \in \{1,2\}$ it holds that \(
    \PE_\pi[\funcAw_{ij}^k] = A_{ij}\) and   \(\PE_\pi[\funcbw_i^k] = b_i\).
\end{assumption}
\begin{assumption}[$p$]
\label{assum:markov:stepsize}
$(\gamma_k)_{k \geq 1}$, $(\beta_k)_{k \geq 1}$ are non-increasing sequences of the form 
\[
\textstyle 
\beta_k = c_{0, \beta} (k+k_0)^{-b}, \quad \gamma_k = c_{0, \gamma} (k+k_0)^{-a} \eqsp, 
\]  
where \(1/2 < a < b < 1\), the ratio  $c_{0,\beta}/c_{0,\gamma}$ is sufficiently small, and the constant $k_0$ satisfies the bound $k_0 \gtrsim p^{4/b}$.
\end{assumption}
\begin{proposition}[Proposition 4 in \cite{butyrin2026gaussian}]
\label{prop:markov:moment_bounds}
Let $p \geq 2 $. Under \Cref{assum:hurwitz,assum:markov:stepsize,assum:UGE_ttsa,assum:aij_bound}, for any $k \geq 0$ it holds that 
\begin{align}
\label{eq:markov:main_ttheta_bound}
\E^{1/p}\big[\|\theta_{k+1} - \theta^\star\|^p\big]&  \lesssim  p^2 \sqrt{\beta_k} \eqsp, \qquad 
\E^{1/p}\big[\|w_{k+1} - w^\star\|^p\big]   \lesssim  p^3 \sqrt{\gamma_k} \eqsp.
\end{align}
\end{proposition}
\begin{remark}
The statement of \Cref{prop:markov:moment_bounds} differs slightly from the original formulation in \cite{butyrin2026gaussian}, as we omit the transient terms. This is without loss of generality, since these terms decay exponentially fast and therefore do not affect the convergence rates.
\end{remark}

\subsection{Proof of Proposition \ref{prop:TTSA-error-bound}}
\label{app:ttsa_rate}

We first verify that the assumptions required for \Cref{prop:markov:moment_bounds} hold in our setting. 

By \eqref{eq:plugin-Schur-stability}, the matrix \(\Delta_\dsieve^{\widehat\pi}\) is positive definite. The matrix \(A_{22,\dsieve}^{\widehat\pi}\), defined in \eqref{eq:TTSA-population-matrix}, is block lower triangular with positive-definite diagonal blocks \(G_{\psi,\dsieve}\), \(G_{\phi,\dsieve}\), and \(G_{\psi,\dsieve}\). Hence all eigenvalues of \(A_{22,\dsieve}^{\widehat\pi}\) are positive, and therefore both \(-\Delta_\dsieve^{\widehat\pi}\) and \(-A_{22,\dsieve}^{\widehat\pi}\) are Hurwitz. Thus, \Cref{assum:hurwitz} is satisfied. Moreover, \Cref{assum:aij_bound} follows directly from \eqref{eq:bounded-features}, the uniform bound \(\|\widehat r \|_\infty\le \lambda\log(\frac{1}{\alpha})\), and the explicit form of the matrices in \eqref{eq:TTSA-population-matrix}. 

 \Cref{assum:UGE}
implies that \( X_n = (S_n', A_n', S_{n+1}')\) is also uniformly geometrically ergodic Markov chain, with
a mixing time differing from \(\tmix\) by at most one step. Consequently, \Cref{assum:UGE_ttsa} follows from \Cref{assum:UGE} and the construction of the sample matrices in \eqref{eq:sample-A}.

Let
\(
    p_\delta
    :=
    \max\{
        2,
        \log\frac{2}{\delta}
    \}.
\)
By \Cref{prop:markov:moment_bounds} and Markov's inequality,
\begin{align}
    \mathbb P\left(
        \|\theta_{N}-\theta^\star_{\dsieve,\widehat\pi}\|
        \gtrsim
        p_\delta^2\sqrt{\beta_N}
    \right)
    \leq
    \frac{\delta}{2},
    \quad
    \mathbb P\left(
        \|w_{N}-w^\star_{\dsieve,\widehat\pi}\|
        \gtrsim
        p_\delta^3\sqrt{\gamma_N}
    \right)
    &\leq
    \frac{\delta}{2}.
\end{align}
Hence, by the union bound, with probability at least \(1-\delta\),
\begin{align}
    \|\theta_{N}-\theta^\star_{\dsieve,\widehat\pi}\|
    \lesssim
    \log^2\frac{2}{\delta}\,
    \sqrt{\beta_N},
    \quad
    \|w_{N}-w^\star_{\dsieve,\widehat\pi}\|
    \lesssim
    \log^3\frac{2}{\delta}\,
    \sqrt{\gamma_N}.
\end{align}
Now set
\(
    a
    =
    1-\frac{1}{\log N}
\) and 
\(
    b
    =
    1-\frac{1}{2\log N}.
\) 
Then, with probability at least \(1-\delta\),
\begin{equation}
\label{eq:ttsa-high-probability-bound}
    \|\theta_{N}-\theta^\star_{\dsieve,\widehat\pi}\|
    +
    \|w_{N}-w^\star_{\dsieve,\widehat\pi}\|
    \lesssim
    \frac{\widetilde C_{\mathrm{SA},\dsieve}\log^3(\frac2\delta)}{\sqrt N}.
\end{equation}
This proves \Cref{prop:TTSA-error-bound}. The constant
\(\widetilde C_{\mathrm{SA},\dsieve}\) is problem-dependent and may depend on the mixing time
\(\tmix\), the effective horizon \(\frac{1}{1-\gamma}\), the uniform feature bound
\(\mathfrak B_{\phi,\dsieve}\), the reward bound \(\mathfrak B_r\), and the inverse
 minimal eigenvalues of the Gram matrices,
\(\lambda_{\min}^{-1}(G_{\phi,\dsieve})\) and
\(\lambda_{\min}^{-1}(G_{\psi,\dsieve})\). Finally, by \eqref{eq:TTSA-reward-output} and
\eqref{eq:Rk-plugin-target},
\[
\begin{aligned}
    \|
        \widehat R_{\dsieve}^{\mathrm{SA}}
        -
        \widehat R_{\dsieve,\widehat\pi}
    \|_{\rho^\star}
    &\leq
  3\mathfrak B_{\phi, \dsieve}
    \left(
        \|\theta_N-\theta_{\dsieve,\widehat\pi}^\star\|
        +
        \|w_N-w_{\dsieve,\widehat\pi}^\star\|
    \right)\le \frac{C_{\mathrm{SA},\dsieve}\log^3(\frac2\delta)}{\sqrt N}\eqsp,
\end{aligned}
\]
where \(C_{\mathrm{SA},\dsieve} = 3\mathfrak B_{\phi, \dsieve}\widetilde C_{\mathrm{SA},\dsieve}\).

\section{Further extensions and interpretations}
\label{sec:extensions}

\paragraph{Projection of rewards onto a given basis.}
Let $\{\psi_1,\ldots,\psi_K\}$ be a fixed collection of features
\(\psi_k:\mathsf S\times\mathsf A\to\mathbb R\), and consider the least--squares problem
\[
    \min_{f\in\mathbb{H}}\;\min_{\alpha_1,\ldots,\alpha_K}\;
    \E_{(s,a)\sim\rho^\star}\!\Big[
        \big(
            R_f(s,a) - \alpha_1 \psi_1(s,a) - \cdots - \alpha_K \psi_K(s,a)
        \big)^2
    \Big].
\]
Let
\(
    \mathcal V:=\operatorname{span}\{\psi_1,\ldots,\psi_K\}
    \subset L^2(\rho^\star).
\)
For a fixed \(f\), the inner minimization is the \(L^2(\rho^\star)\)-orthogonal
projection of \(R_f\) onto \(\mathcal V\). Hence, the problem is equivalent to
\begin{equation}
\label{eq:extension_res}
    \min_{f\in\mathbb H}
    \big\|
        R_f-\Pi_{\mathcal V}R_f
    \big\|_{L^2(\rho^\star)}^2.
\end{equation} 

Geometrically, this criterion selects a reward with the smallest
\(L_2(\rho^\star)\)-distance to the prescribed feature space
\(\mathcal V\), while
the residual \eqref{eq:extension_res} quantifies the part
of the reward that cannot be captured by the chosen features. 
The features may encode prior structural knowledge of the environment believed to be relevant for the reward. This
provides a direct mechanism for incorporating such prior information into the
reward-recovery procedure.

\paragraph{Connection to the Bradley--Terry model.}

\(\ERIRL\) naturally leads to a probabilistic choice rule of the 
form \eqref{eq:softmax}
where $Q^\star$ is the soft (entropy--regularized) action--value function.
This has exactly the same mathematical form as the
\emph{Bradley--Terry--Luce} (BTL) model (\cite{BradleyTerry}) for discrete choice and pairwise
comparisons.  
In the BTL model, each item $i$ has a latent ``utility'' $u_i$, and the probability 
that $i$ is chosen over a set of alternatives $\mathcal{I}$ is
\[
    P(i)
    \;=\;
    \frac{\exp(u_i)}{\sum_{j\in\mathcal{I}} \exp(u_j)}.
\]
Thus, conditional on a state $s$, \ERIRL\, induces a BTL choice model over actions
with utilities
\[
    u_a(s)
    \;=\;
    \tfrac{1}{\lambda} Q^\star(s,a).
\]
Note that  the \ERIRL\, soft $Q$--values $Q^\star(s,a)$ encode these same preferences
    up to a temperature parameter $\lambda$
    and the optimal policy $\pi^\star(\cdot\mid s)$ is the BTL choice
    distribution over actions induced by the utilities $\{u_a(s)\}$.
From the IRL perspective, we observe actions chosen by the expert and must infer
the underlying utilities $u_a(s)$ (or equivalently $Q^\star(s,a)$ and thus the
reward).  
This is precisely analogous to estimating latent utility parameters in the
Bradley--Terry model from observed choices.
More deeply, the \ERIRL\, likelihood
is the log-likelihood of a sequence of contextual BTL choices, where the
context is the state $S_t$.  
\ERIRL\, therefore amounts to a contextual Bradley--Terry inversion 
problem: from repeated state--dependent choices, infer the underlying 
(preference--encoded) reward function. Finally, $r_a(s)=\lambda\log\pi^\star(a\mid s)$ is the centered, policy-identifiable part of the soft utility: it differs from $Q^\star(s,a)$ by the state-dependent offset $V^\star(s)$. The least--squares reward fitting step can therefore be interpreted as 
recovering a reward function whose induced utilities best explain the observed 
Bradley--Terry--style preference probabilities.
In this sense, \ERIRL\, can be viewed as a structured generalization of
the Bradley--Terry model in which utilities are constrained by Bellman
consistency and derived from an underlying reward function over state--action 
pairs.

\end{document}